\documentclass[letterpaper, 10 pt, conference]{ieeeconf}  

\IEEEoverridecommandlockouts                              

\usepackage{graphicx} 
\usepackage{subcaption}
\usepackage{textcomp}
\usepackage{array}
\usepackage{booktabs}
\usepackage[dvipsnames]{xcolor}
\usepackage{tikz}
\usepackage[export]{adjustbox}
\usepackage{url}
\usepackage{multirow}
\usepackage{soul}
\usepackage{hyperref}
\usepackage{siunitx}

\long\def\invis#1{}
\newcommand\sect[1]{Section~\ref{#1}}
\newcommand\tab[1]{Table~\ref{#1}}
\newcommand\fig[1]{Figure~\ref{#1}}
\newcommand\eq[1]{Equation~(\ref{#1})}

\newcommand\copyrighttext{%
  \footnotesize \textcopyright 2019 IEEE. Personal use of this material is permitted.
  Permission from IEEE must be obtained for all other uses, in any current or future
  media, including reprinting/republishing this material for advertising or promotional
  purposes, creating new collective works, for resale or redistribution to servers or
  lists, or reuse of any copyrighted component of this work in other works.
  DOI: \href{https://doi.org/10.1109/IROS40897.2019.8967557}{10.1109/IROS40897.2019.8967557}}
\newcommand\copyrightnotice{%
\begin{tikzpicture}[remember picture,overlay]
\node[anchor=south,yshift=10pt] at (current page.south) {\fbox{\parbox{\dimexpr\textwidth-\fboxsep-\fboxrule\relax}{\copyrighttext}}};
\end{tikzpicture}%
}

\title{\LARGE \bf
Real-time Model-based Image Color Correction for Underwater Robots
}

\author{Monika Roznere and Alberto Quattrini Li
\thanks{The authors are with Department of Computer Science,
        Dartmouth College, Hanover, NH USA
        {\tt\small \{monika.roznere.gr, alberto.quattrini.li\}@dartmouth.edu}}%
}

\begin{document}

\maketitle
\copyrightnotice
\thispagestyle{empty}
\pagestyle{empty}

\begin{abstract}
Recently, a new underwater imaging formation model presented that the coefficients related to the direct and backscatter transmission signals are dependent on the type of water, camera specifications, water depth, and imaging range. This paper proposes an underwater color correction method that integrates this new model on an underwater robot, using information from a pressure depth sensor for water depth and a visual odometry system for estimating scene distance. Experiments were performed with and without a color chart over coral reefs and a shipwreck in the Caribbean. We demonstrate the performance of our proposed method by comparing it with other statistic-, physic-, and learning-based color correction methods. Applications for our proposed method include improved 3D reconstruction and more robust underwater robot navigation.
\end{abstract}

\section{INTRODUCTION}
Over the last decade, scientists have increasingly been using cameras carried by hand or on-board of Remotely Operated Vehicles (ROVs) or Autonomous Underwater Vehicles (AUVs)~\cite{bryson2016true,iscar2018towards} to collect imagery data in underwater environments. Several applications benefit from such a technology, including cave mapping \cite{weidner2017underwater}, coral reef monitoring~\cite{quattrinili2016iser-mr}, and preserving archaeological sites~\cite{henderson2013mapping}. Scientists need high quality images to be able to, for example, reconstruct realistic underwater structures or evaluate the health of coral reefs~\cite{skinner2017automatic}. In addition, robots need high quality images to reliably navigate underwater.

Compared to in-air images, underwater images are more affected by blurriness, distortions, and light attenuation. Furthermore, bodies of water have different optical properties, depending on location, season, weather, and quality of marine life, especially of phytoplankton. As light propagates underwater, it interacts with suspended particles, causing rays of light to be \emph{absorbed} or \emph{scattered}. This combined phenomena of absorption and scattering is called \emph{attenuation}. Light attenuation is also related to wavelength, as red light degrades more quickly than blue light. Hence, the majority of underwater images lack red color. The current underwater imaging formation model tries to comply with the laws of light propagation in underwater. Last year, a new underwater imaging formation model was proposed by~\cite{Akkaynak2018}, showing that the underwater image model depends on more complex factors, including water type (classified by the ten Jerlov's water types \cite{Solonenko2015}), camera specifications, water depth, and imaging range. 

\begin{figure}[t]
\begin{minipage}[b]{0.45\linewidth}
\centering
\includegraphics[width=\textwidth]{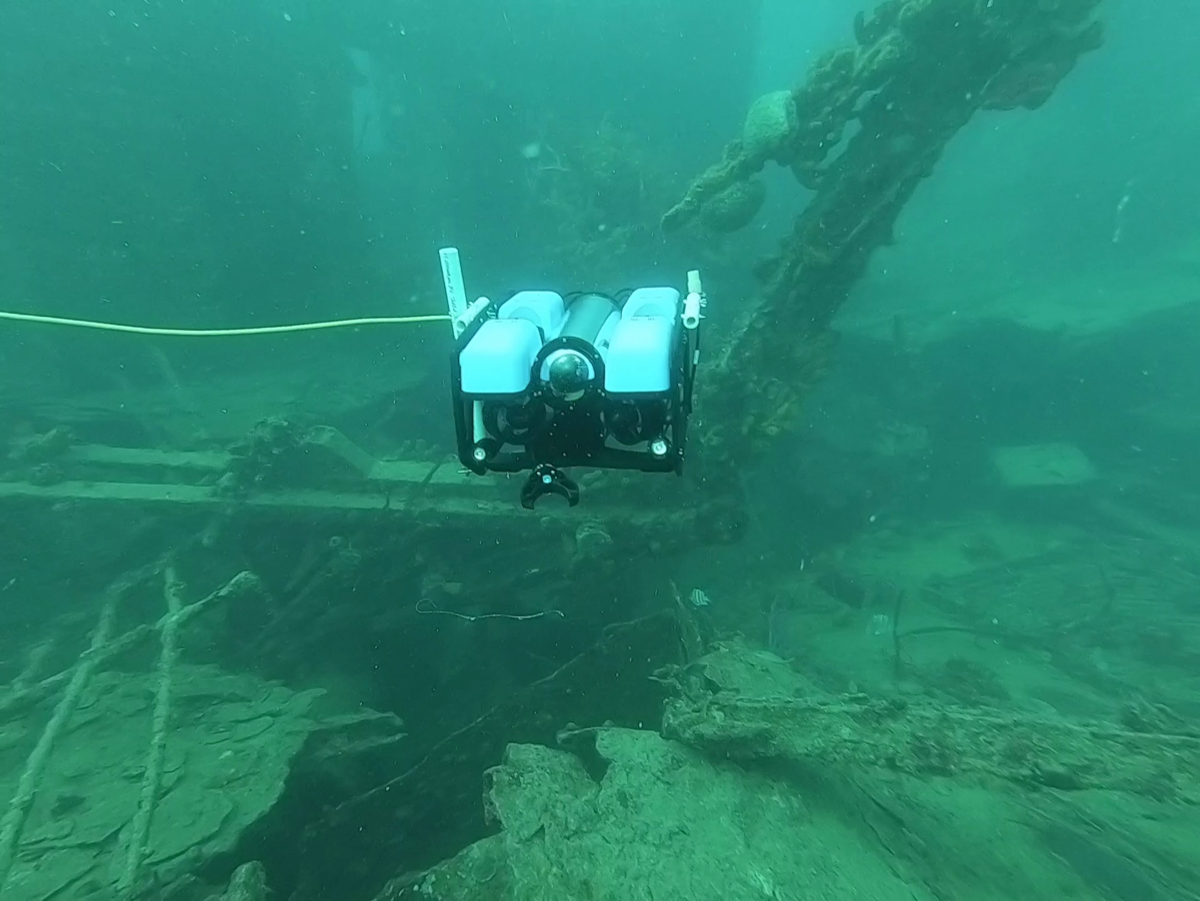}
\label{fig:figure1}
\end{minipage}
\hspace{0.2cm}
\begin{minipage}[b]{0.48\linewidth}
\centering
\includegraphics[width=\textwidth]{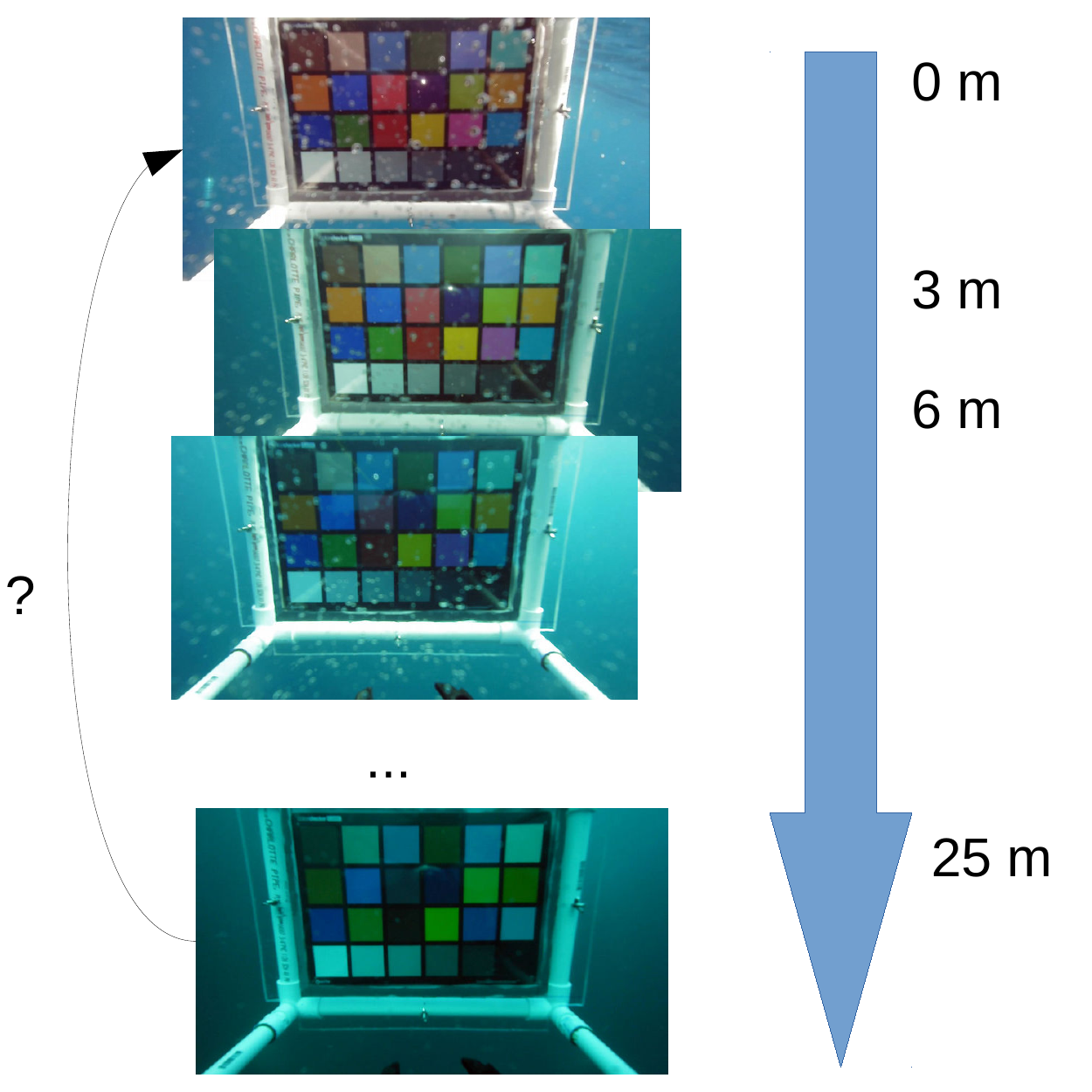}
\end{minipage}
\caption{BlueROV2 with a camera over a shipwreck, and color degradation of a color chart over depth. }
\label{fig:problem}
\end{figure}

This paper addresses the problem of underwater color correction and implements a new method based on the recently revised underwater imaging formation model.
The main contributions of the paper are:
\begin{itemize}
    \item a novel color correction method based on the new underwater imaging formation model;
    \item direct integration of the method within a robot (\fig{fig:problem} shows the robot used for experiments and sample data collected at different depths), using the pressure sensor for water depth and a visual odometry method to estimate object distance;
    \item systematic experiments in Barbados to observe color changes at different water depths and object distances.
\end{itemize}

The proposed method is compared with other state-of-the-art methods, including Contrast Limited Adaptive Histogram Equalization (CLAHE)~\cite{pizer1987adaptive}, white-balancing using gray world assumption~\cite{buchsbaum1980spatial}, Underwater Generative Adversarial Network (UGAN)~\cite{fabbri2018enhancing}, and Fusion method~\cite{ancuti2012enhancing}. Through quantitative and qualitative evaluations, the results present that our proposed method has high color accuracy over great changes of depth and viewing distance, as well as high color consistency. We lastly demonstrate that our proposed method is compatible with other robotic tasks, such as visual odometry with ORB SLAM2 \cite{Mur2015}. 

The paper is structured as follows. The next section presents image color correction methods, mainly focusing in the underwater domain. \sect{sec:approach} describes the proposed method. \sect{sec:results} and \sect{sec:discussion} present and discuss the experimental results. Finally, \sect{sec:conclusion} concludes the paper.

\section{RELATED WORK}
Methods addressing underwater image color correction can be classified in three main categories: \emph{statistics-based methods}, \emph{physics-based methods}, and \emph{learning-based methods}. Here, we describe a set of representatives from each category and refer the interested reader to \cite{schettini2010underwater} for a survey.

Statistics-based methods, such as local histogram equalization \cite{garcia2002way}, automatic color equalization~\cite{chambah2003underwater}, gray world assumption~\cite{buchsbaum1980spatial}, adjust the distribution of pixel intensities, possibly in different color spaces and by shifting the mean or higher order statistics.  \cite{bianco2015new} proposes a method that works in the LAB color space under a white world assumption.
While, \cite{ancuti2012enhancing,ancuti2018color} develop a multi-scale fusion technique that merges two color corrected images to be blended in a weight map that defines the desired level of correction. \cite{fu2014retinex} proposes a method that decomposes reflectance and illumination and post-processes each with a histogram-based technique.
While this family of methods improves imaging contrast and luminosity, they do not always correct the color according to its ground truth color.

Other methods use the underwater imaging formation model. This is typically composed of the direct and the backscatter signal, which is the light that directly goes from the object to the camera and the light that scatters due to particles present in the water before reaching to the camera, respectively, to restore underwater images.
\cite{chiang2012underwater} estimates the depth (distance from the object) by using the dark channel prior~\cite{he2011single} and by dehazing. Then they compensate the wavelengths of each channel according to the underwater imaging formation model.
With the same model, \cite{lu2013underwater} proposes a simple prior that is calculated from the difference in attenuation among the red color channels, which then can be used to estimate the transmission map and noise filter. On the other hand, \cite{carlevaris2010initial} exploits the attenuation differences between the RGB channels in water to estimate the depth of the scene, and then uses the underwater imaging formation model to correct the image.
Last year, \cite{Akkaynak2018} proposed a new underwater imaging formation model, where the coefficients for the direct and the backscatter signals are different and depend on parameters, including water depth, object distance, and water type. This requires modifications to the previous methods.
Based on how light propagates and attenuates underwater, researchers proposed the use of polarizers~\cite{schechner2004clear,schechner2005recovery}, spectrometers~\cite{aahlen2007application}, Doppler Velocity Loggers (DVL) and beam patterns~\cite{kaeli2011improving}, and color charts~\cite{Skaff2008} to estimate the physical parameters of the underwater imaging formation model, as well as the spectral properties of the objects in the scene. 
Adaptive artificial illumination has also been implemented to correctly compensate for color loss~\cite{vasilescu2011color}.

Learning based methods rely on underwater training data sets. 
The Markov Random Field learning based model can be implemented to learn the relationship between a corrected image and its corresponding attenuated image \cite{Torres2005}. However, it heavily relies on the availability of ground truth images. Similarly, \cite{protasiuk2019local} combines local color mapping, between known colors from a target, and color transfer, between colors and a reference image, to apply an affine transformation to the underwater image.
However, one of the challenges for underwater color correction is that ground truth is difficult to obtain. 
In the last few years, several methods based on a Generative Adversarial Network (GAN) have been proposed. 
WaterGAN~\cite{li2018watergan} requires a training set composed of in-air images and a corresponding depth map to generate underwater images. Their method then color corrects underwater images by using two convolutional neural networks, which produce a depth map and the color corrected image.
Other GAN-based methods have been proposed to translate an image directly to another image~\cite{fabbri2018enhancing,li2018emerging}, without requiring depth information.

Our proposed color correction method does not require any training data set and is a physics-based method, inspired by the new imaging formation model proposed by \cite{Akkaynak2018}.

\section{APPROACH}\label{sec:approach}
\begin{figure}
    \centering
    \includegraphics[width=\columnwidth]{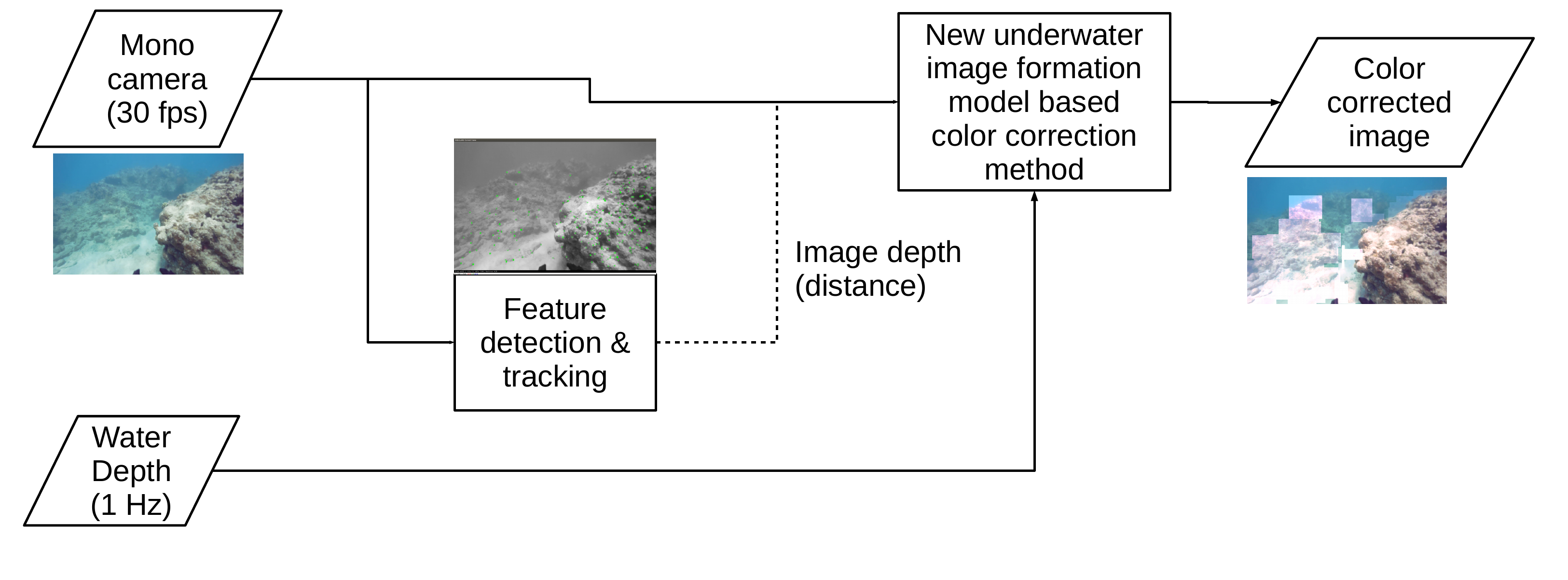}
    \caption{High-level pipeline of our proposed approach.}
    \label{fig:pipeline}
\end{figure}

The main idea of our approach is to color correct the image using the new underwater imaging formation model proposed by~\cite{Akkaynak2018} that assumes that the attenuation coefficients for backscatter and direct signal are different. We propose a system that automatically derives the two attenuation coefficients necessary to solve the imaging formation model equation, so that images can be color corrected directly on-board an underwater robot. The high-level pipeline is shown in \fig{fig:pipeline} and is detailed in the following.

The equation of the underwater imaging formation model~\cite{Akkaynak2018} expresses the raw image $I_c$, where $c$ represents each of the $\textit{RGB}$ color channels, to be:
\begin{equation}
    I_c = J_c e^{-\beta_c^D(\mathbf{v}_D)z} + B_c^\infty (1 - e^{-\beta_c^B(\mathbf{v}_B) z}),
    \label{eq:underwater-image-formation-model}
\end{equation}

\noindent where $J_c$ is the unattenuated image, $z$ is the imaging range, $B_c^\infty$ the wideband veiling light, and $\beta_c^D$ and $\beta_c^B$ are the attenuation coefficients related to direct signal and backscatter, respectively. Note that in the new underwater imaging formation model, the attenuation coefficients depend on $\mathbf{v}_D=[z,\rho,E,S_c,\beta]$ and $\mathbf{v}_B=[E,S_c,b,\beta]$, where $\rho$ is the reflectance spectrum of the object, $E$ is the ambient light, $S_c$ is the spectral response of the camera, and $\beta=a+b$ is the beam attenuation coefficient, with $a$ and $b$ being the beam absorption and scattering coefficient. Wideband veiling light $B_c^\infty$ ~\cite{Akkaynak2018} for a color channel $c$ is calculated as follows: 

\begin{equation}
    B_c^\infty = \frac{1}{k} \int_{\lambda1}^{\lambda2} S_c(\lambda) \frac{b_c E_c(d,\lambda)}{\beta_c} d\lambda ,
\end{equation}

\noindent where $k$ is a scalar directing image exposure and $\lambda$ is the wavelength. $E(d,\lambda)$, the ambient light at a given wavelength $\lambda$, at depth $d$, is 

\begin{equation}
    E(d,\lambda) = E_0 \frac{K_d(\lambda)}{d},
\end{equation}

\noindent where $E_0$ is the ambient light at the surface and $K_d$ is the diffuse attenuation coefficient. The coefficients $a$, $b$, and $K_d$ depend on the type of water, as defined by Jerlov \cite{Solonenko2015}, and can be derived from the current charts.

From previous testing, we observed that the reflectance spectrum of the object $\rho$ provides negligible impact to the overall calculations. One possible explanation is the increase of diffuse reflection seen in rough material, commonly found in underwater reliefs. For simplicity, we assume $\rho$ to be 1. Similarly, we assume the ambient light at the surface $E_0$ to be 1. $S_c$ can be easily obtained through technical documents (e.g., for the camera used in our experiments \cite{sonyimx322lqjc}). The water depth $d$ is measured with a pressure depth sensor. Distance $z$ is either a known scalar value or estimated through a visual odometry package -- ORB SLAM2~\cite{Mur2015} -- that provides a sparse estimate for each image.

In cases where the veiling light cannot be calculated or the coefficients are unavailable, the veiling light can be assumed to be the average pixel value in the distance, or the background color. When water conditions are poor or prior knowledge is unreliable, this approach can be implemented.

\begin{figure}
    \centering
    \includegraphics[width=0.9\columnwidth]{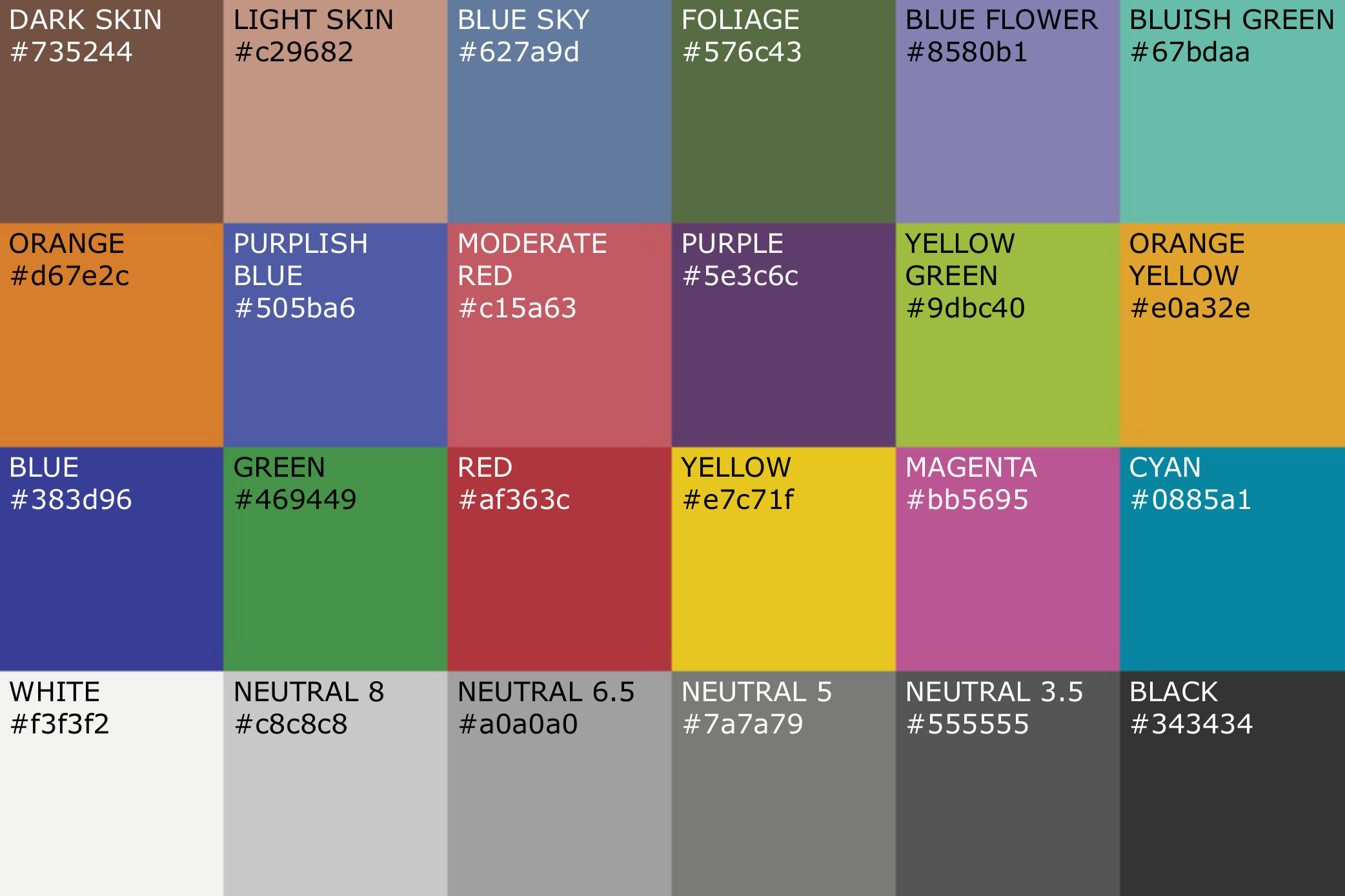}
    \caption{Simulated X-Rite ColorChecker\textsuperscript{\tiny\textregistered} Classic.}
    \label{fig:simulated-color-chart}
\end{figure}

If $J_c$ is known, for instance when a color chart is utilized, $\beta_c^D$ and $\beta_c^B$ can be estimated by taking color pixel samples from two color patches, see Figure \ref{fig:simulated-color-chart}, of the same image. We estimate using white and black patches. Our motivation behind this is that the white and black colors represent the wide color spectrum, thus also representing the majority of the color degradation that occurs.

As shown in the pipeline, the process receives the current image, an approximate depth reading, and image depth sparse map. The attenuation coefficients are estimated using information from the image scene. Then, \eq{eq:underwater-image-formation-model} is used to solve for $J_c$ the unattenuated image, also known as the corrected image. Lastly, the corrected image is published for further use by other tasks, such as scene reconstruction.

\begin{figure}
    \centering
    \includegraphics[width=.9\columnwidth]{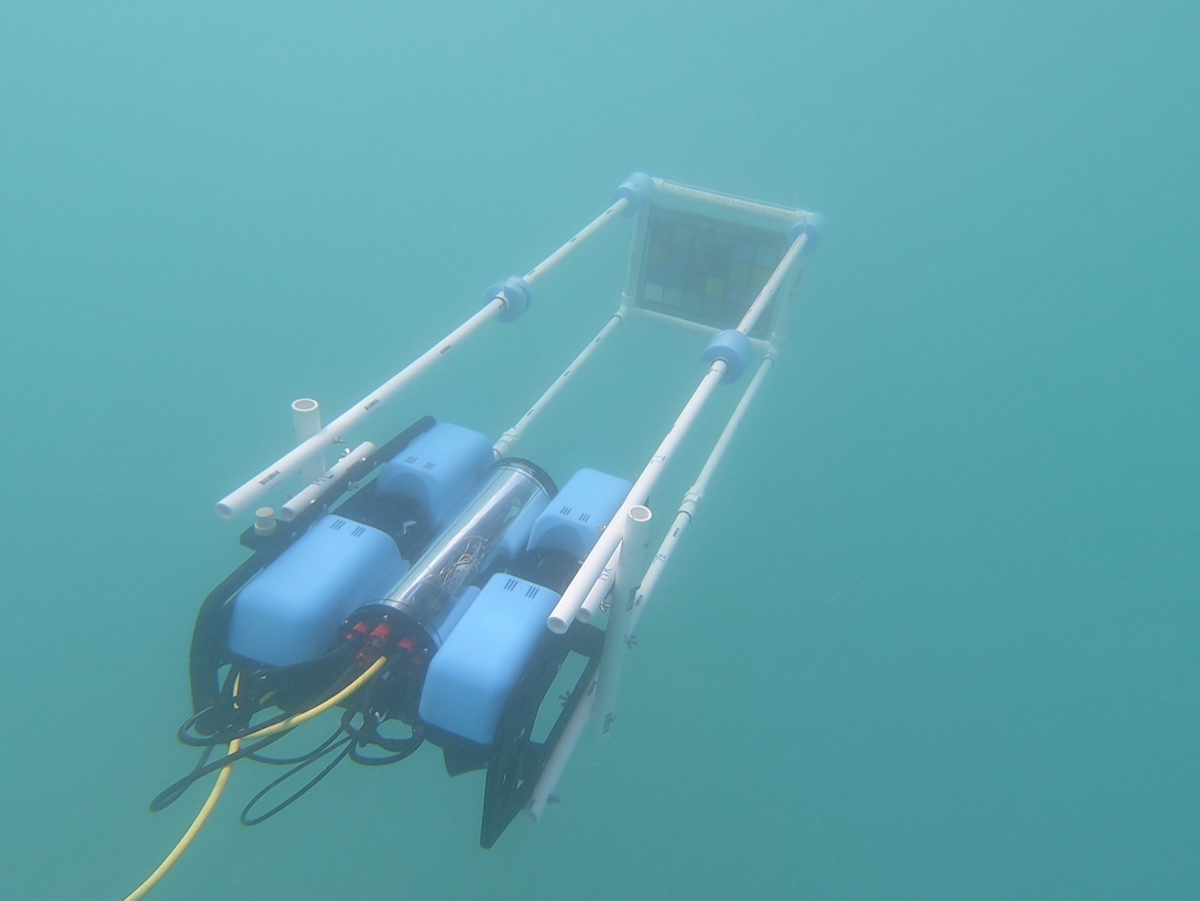}
    \caption{BlueROV2 with color chart attached.}
    \label{fig:robot-with-color-chart}
\end{figure}

\section{EXPERIMENTAL RESULTS}\label{sec:results}

All experiments and data collection were performed at different locations in the Caribbean Sea off the coast of Barbados. We deployed the BlueROV2 and executed color enhancement methods on the images taken by its installed Sony IMX322LQJ-C camera~\cite{sonyimx322lqjc}. Our proposed method has been implemented in Python and integrated with ROS~\cite{quigley2009ros}\footnote{The code will be made opensource on our lab website \url{https://rlab.cs.dartmouth.edu}}. We ran our code on a laptop with Ubuntu 16.04 equipped with an Intel i7, 16 GB of RAM, used together with the ROV. Our first set of experiments evaluates color accuracy and color consistency. A color chart was used as ground truth, as shown in Figure \ref{fig:simulated-color-chart}. Our second set of experiments demonstrates the capability of color correction using the information provided from a SLAM system. We perform method comparisons to validate that the proposed method is applicable to real-time robotic tasks.

\subsection{Color Accuracy and Consistency}

\newcolumntype{M}[1]{>{\centering\arraybackslash}m{#1}}
\begin{table*}[htb]
  \centering
  \begin{tabular}{c c c c c c c c}
    \toprule
    depth & a & b & c & d & e & f & g \\
    \midrule
    \SI{3.26}{\m} & \includegraphics[width=.10\textwidth, valign=c]{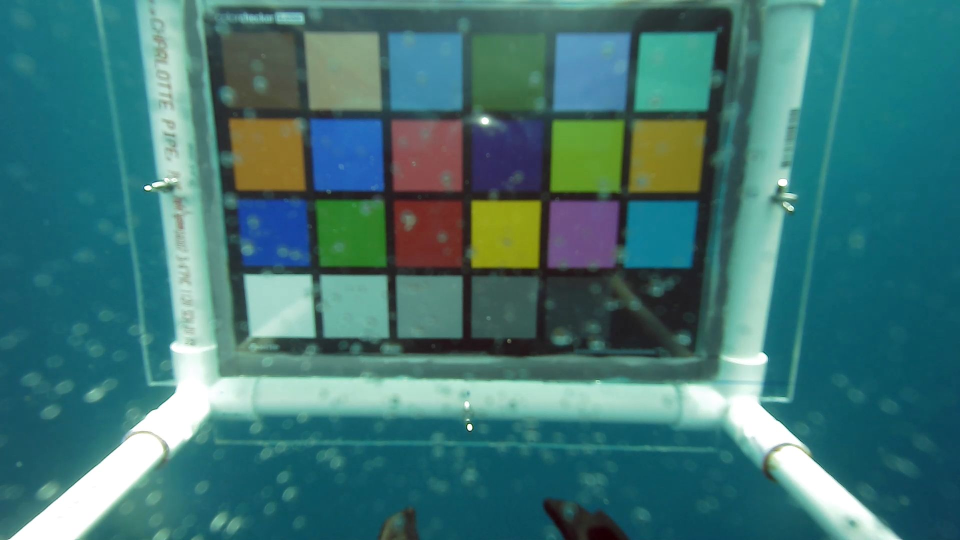}
           & \includegraphics[width=.10\textwidth, valign=c]{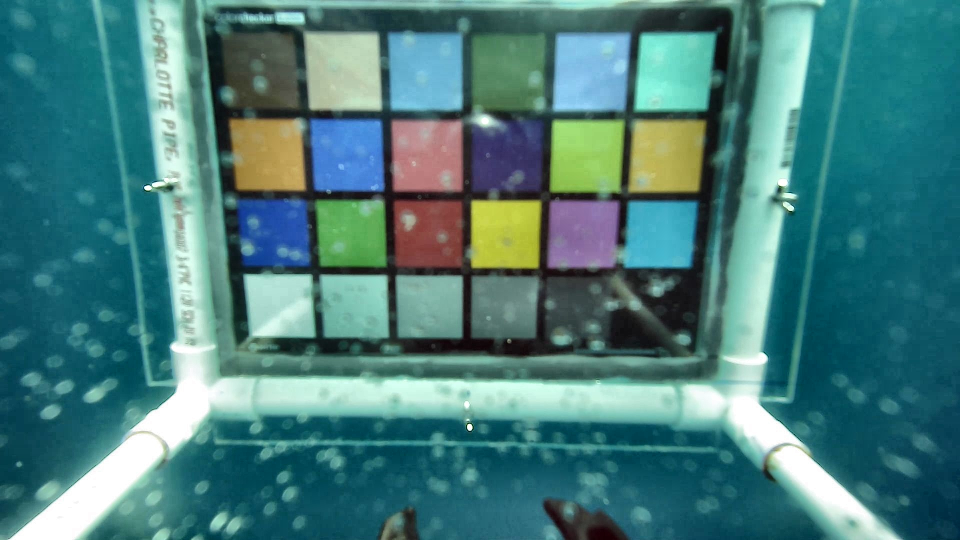}
           & \includegraphics[width=.10\textwidth, valign=c]{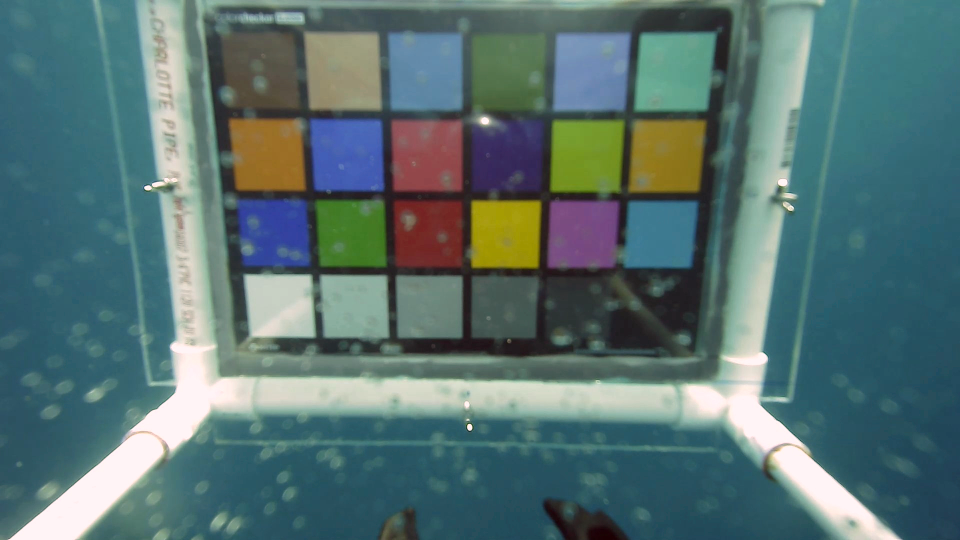}
           & \includegraphics[width=.10\textwidth, valign=c]{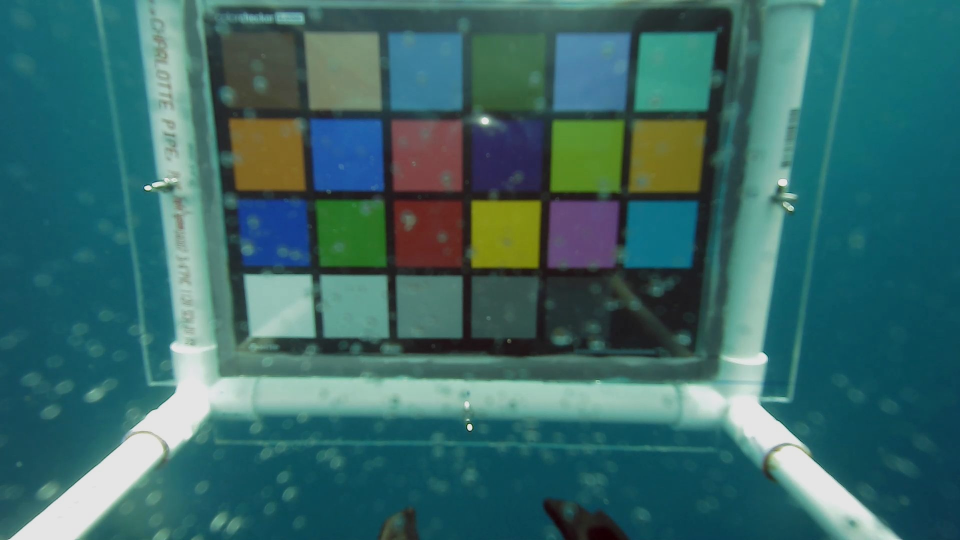}
           & \includegraphics[width=.10\textwidth, valign=c]{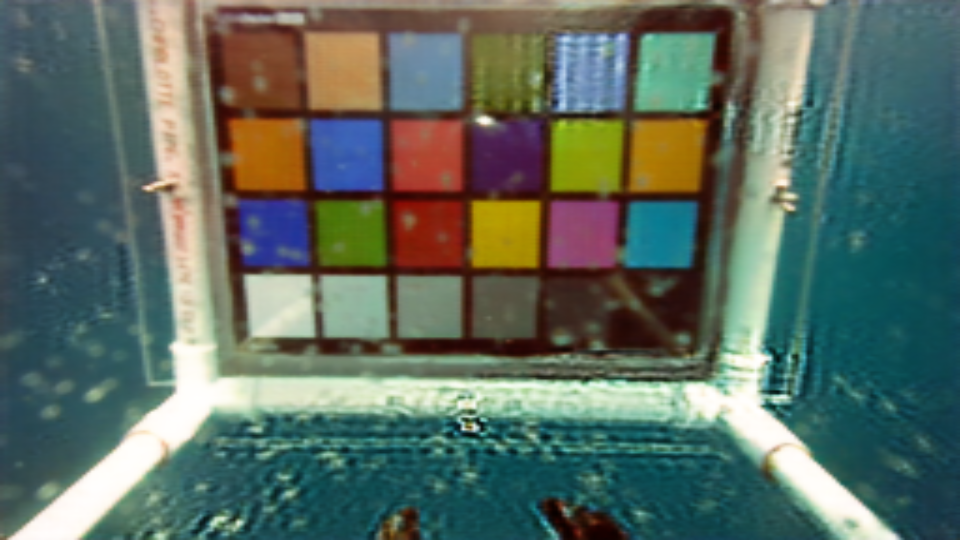}
           & \includegraphics[width=.10\textwidth, valign=c]{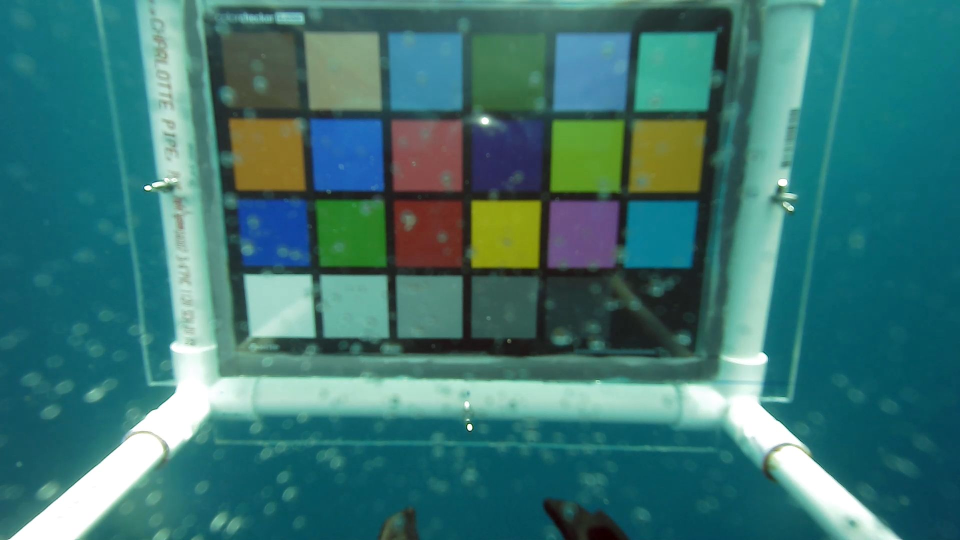}
           & \includegraphics[width=.10\textwidth, valign=c]{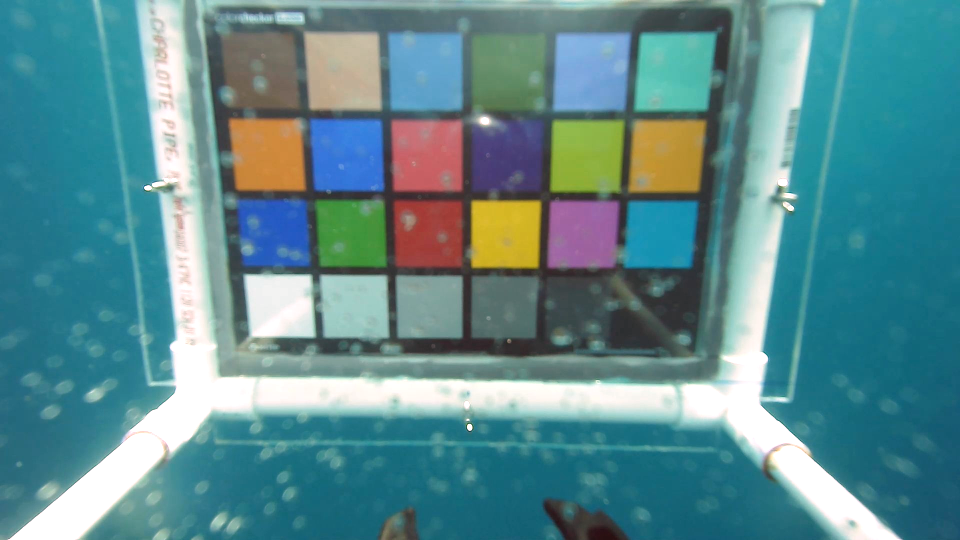} \\
    \SI{6.06}{\m} & \includegraphics[width=.10\textwidth, valign=c]{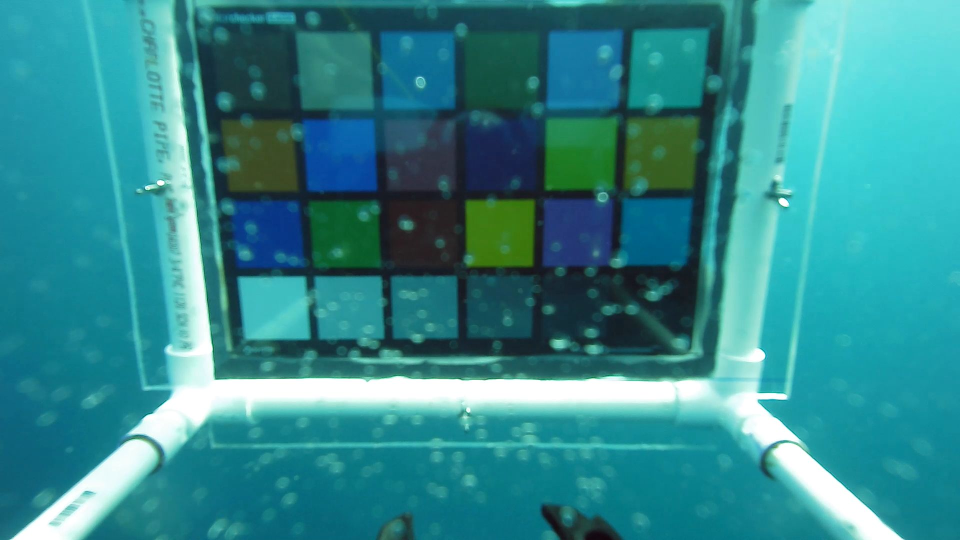}
           & \includegraphics[width=.10\textwidth, valign=c]{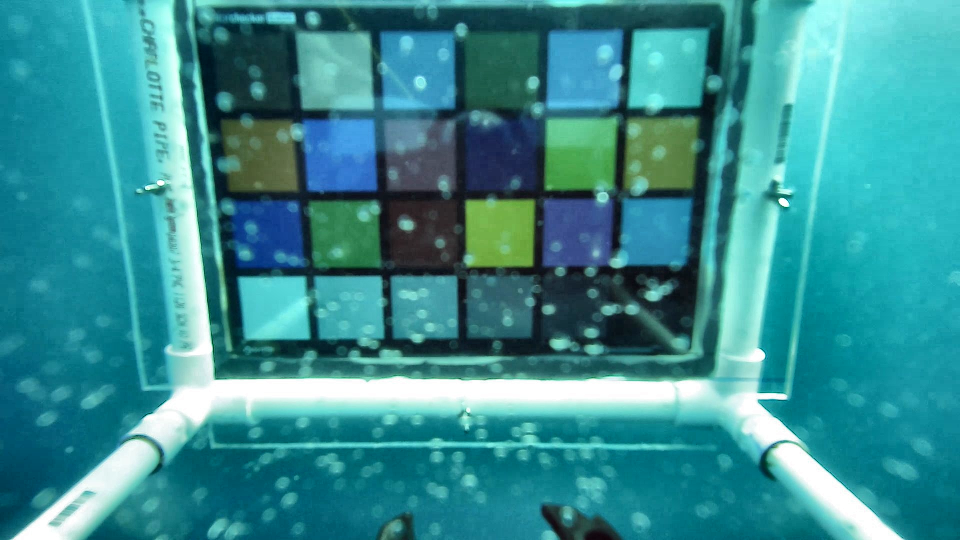}
           & \includegraphics[width=.10\textwidth, valign=c]{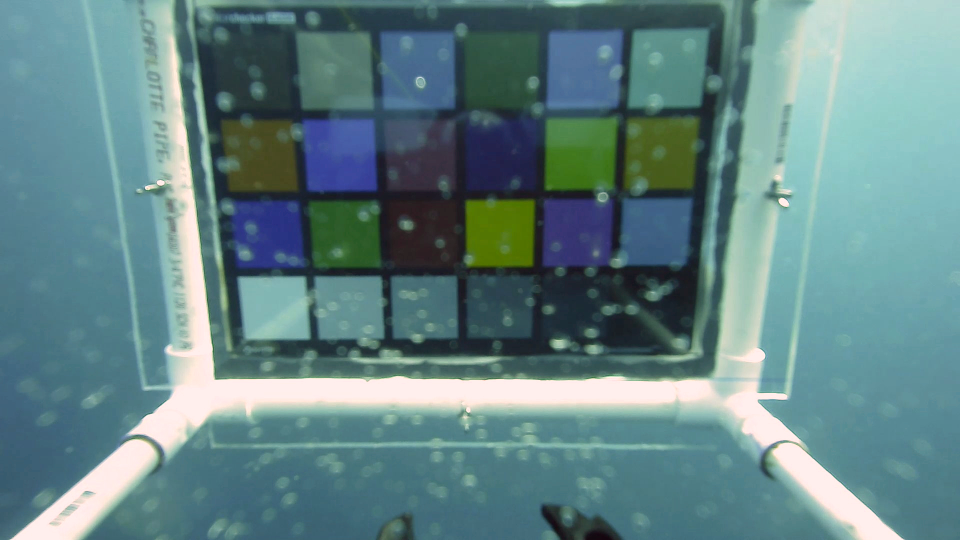}
           & \includegraphics[width=.10\textwidth, valign=c]{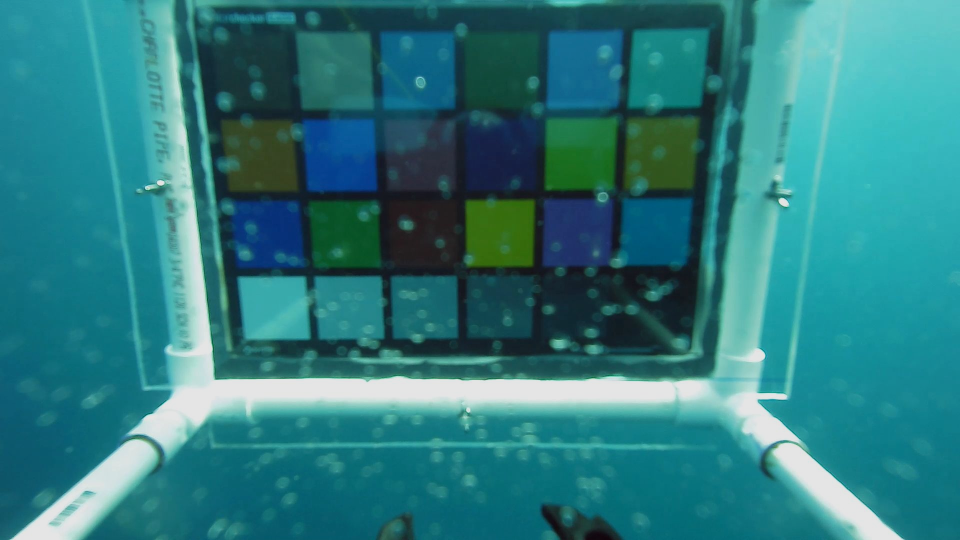}
           & \includegraphics[width=.10\textwidth, valign=c]{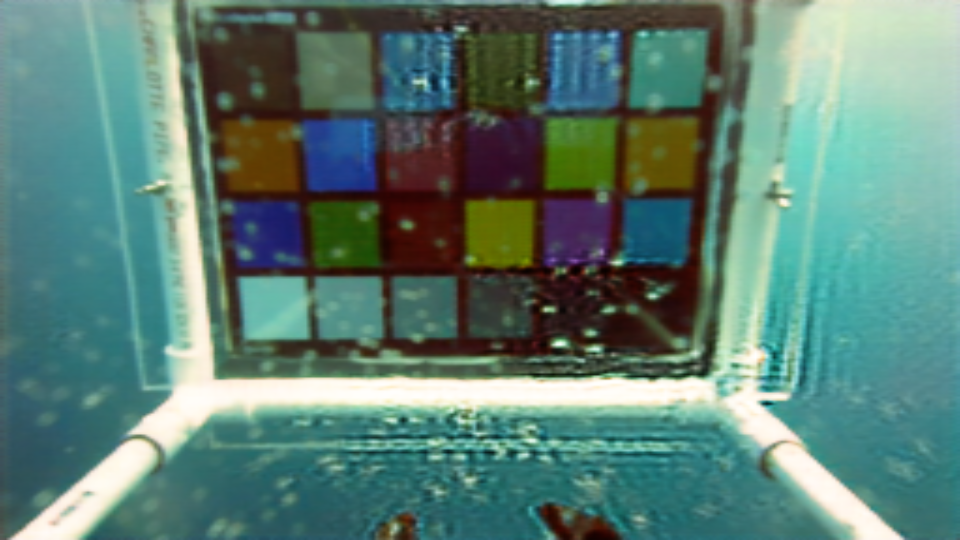}
           & \includegraphics[width=.10\textwidth, valign=c]{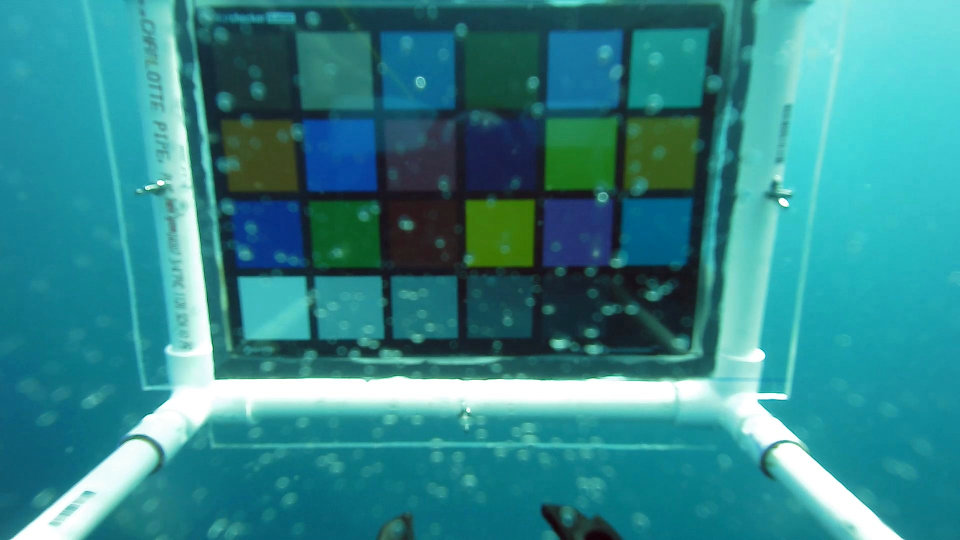}
           & \includegraphics[width=.10\textwidth, valign=c]{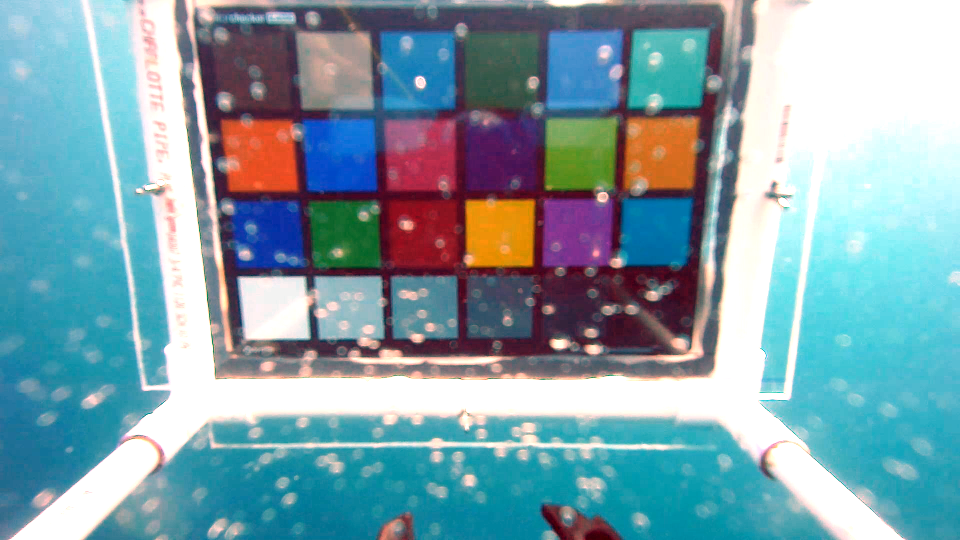} \\
    \SI{8.98}{\m} & \includegraphics[width=.10\textwidth, valign=c]{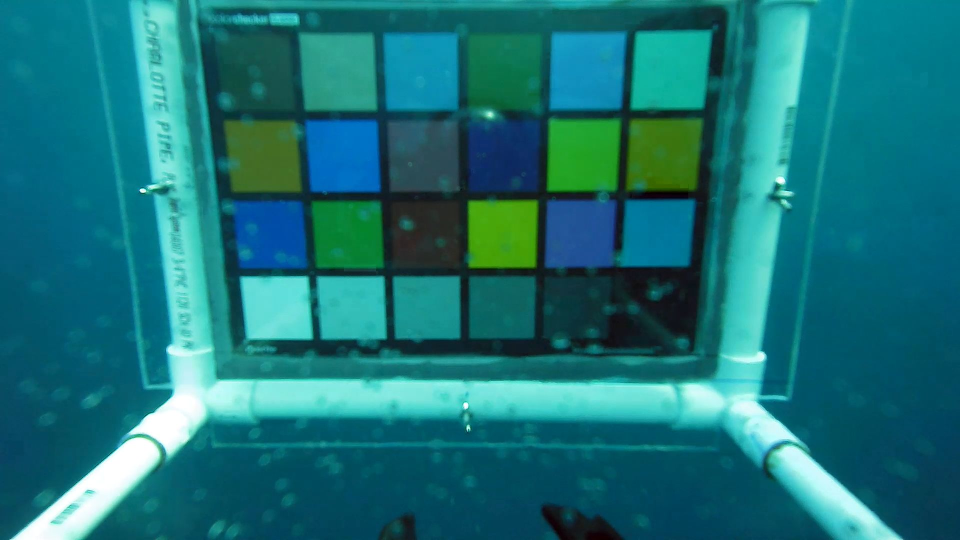}
           & \includegraphics[width=.10\textwidth, valign=c]{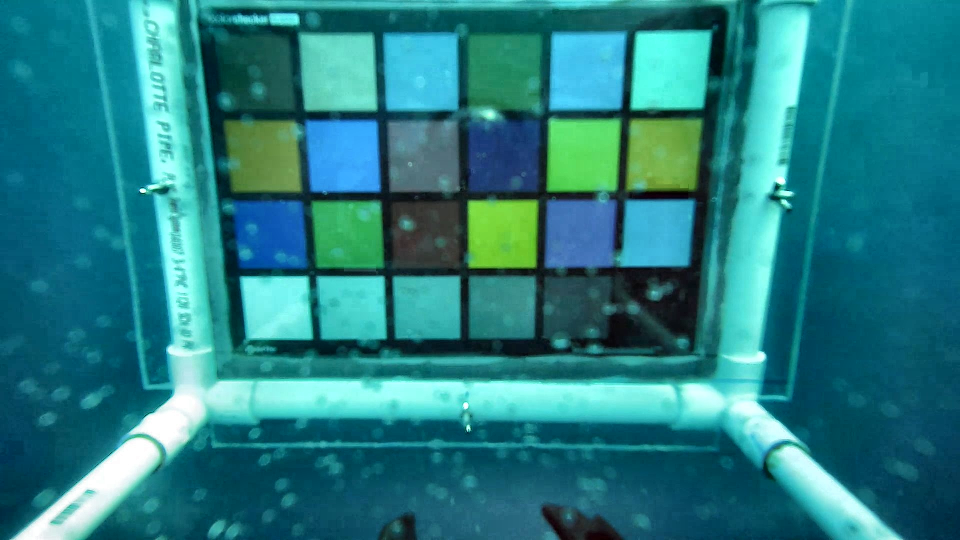}
           & \includegraphics[width=.10\textwidth, valign=c]{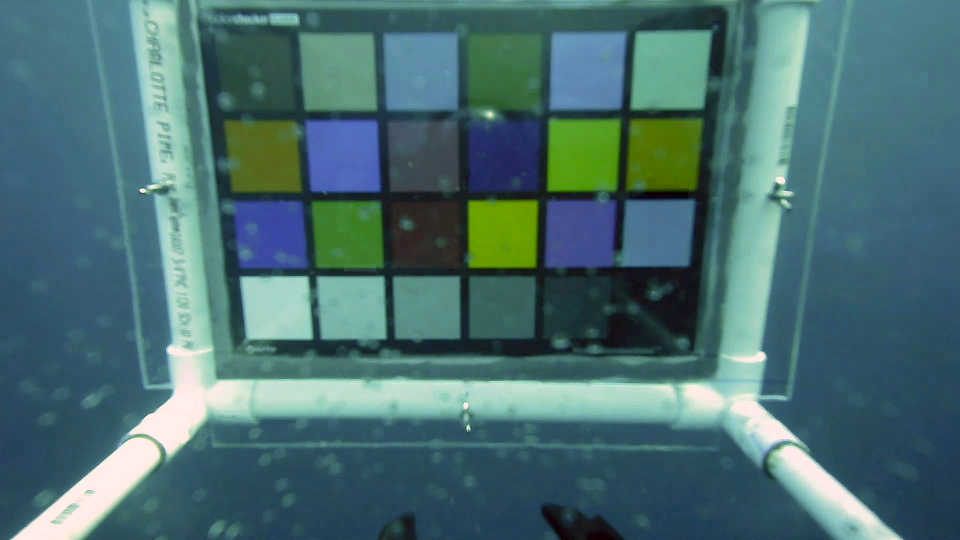}
           & \includegraphics[width=.10\textwidth, valign=c]{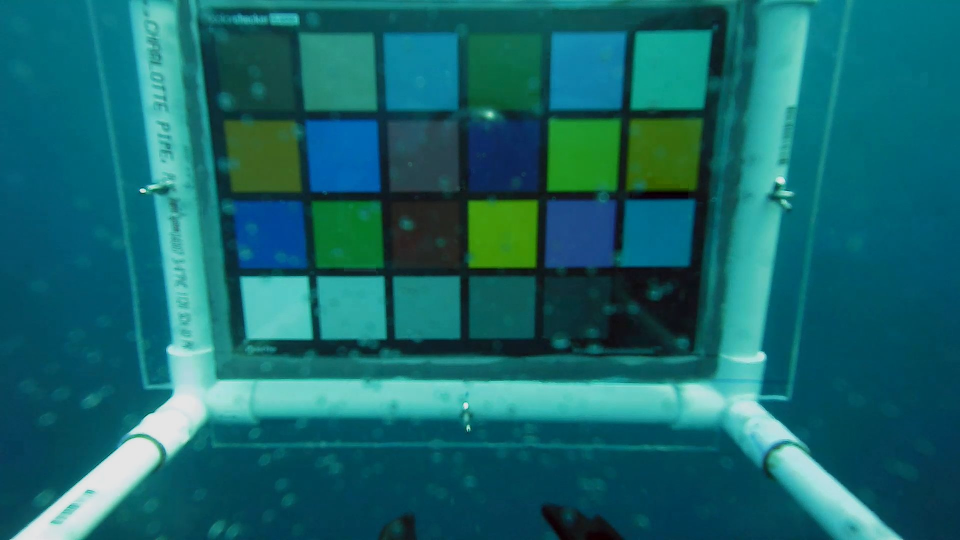}
           & \includegraphics[width=.10\textwidth, valign=c]{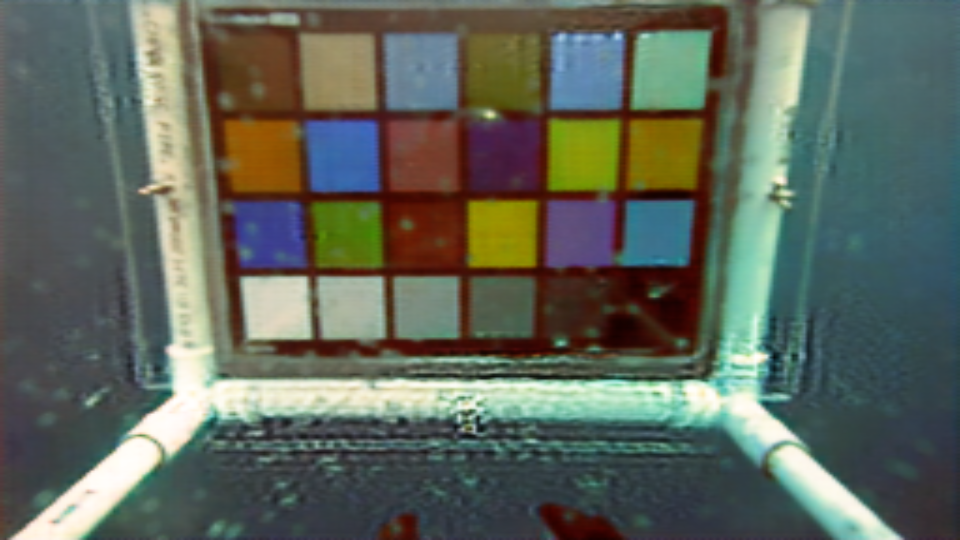}
           & \includegraphics[width=.10\textwidth, valign=c]{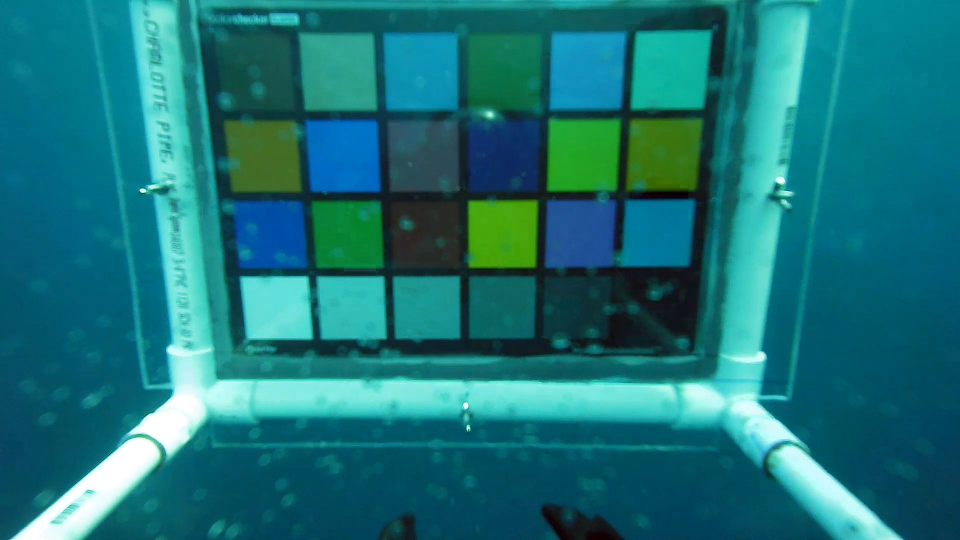}
           & \includegraphics[width=.10\textwidth, valign=c]{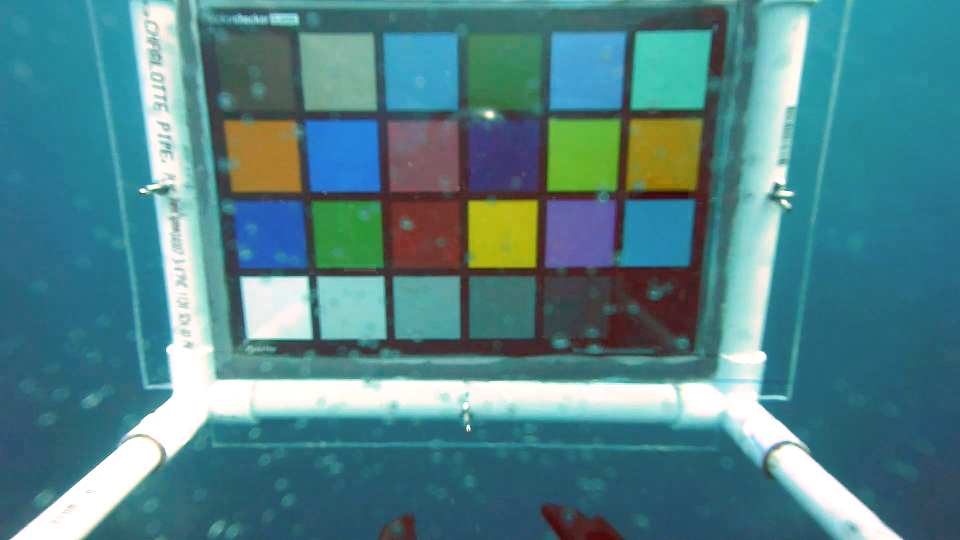} \\
    \SI{12.25}{\m} & \includegraphics[width=.10\textwidth, valign=c]{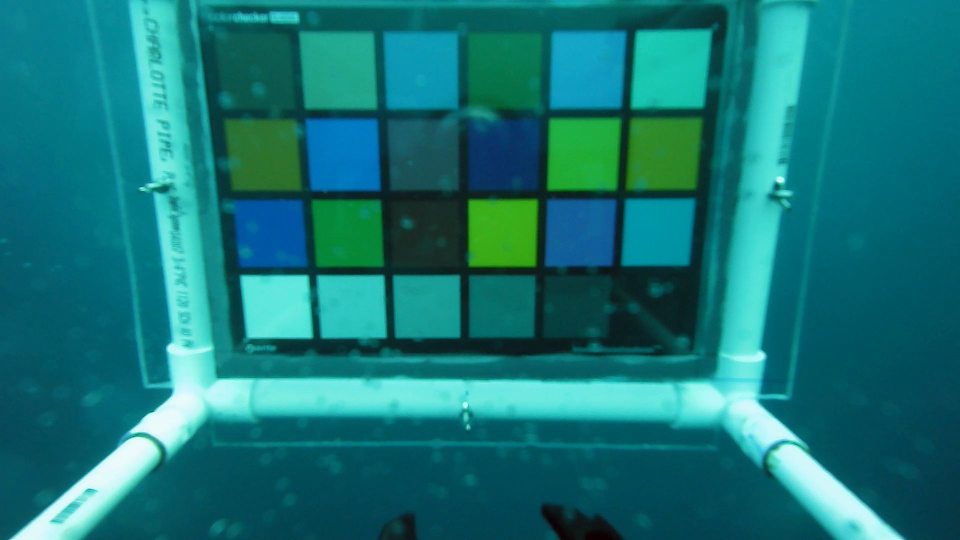}
            & \includegraphics[width=.10\textwidth, valign=c]{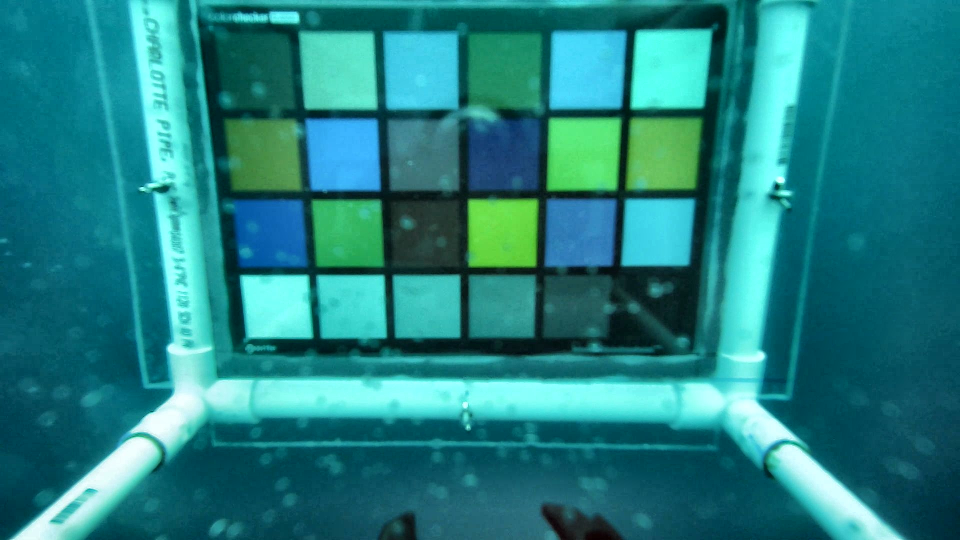}
            & \includegraphics[width=.10\textwidth, valign=c]{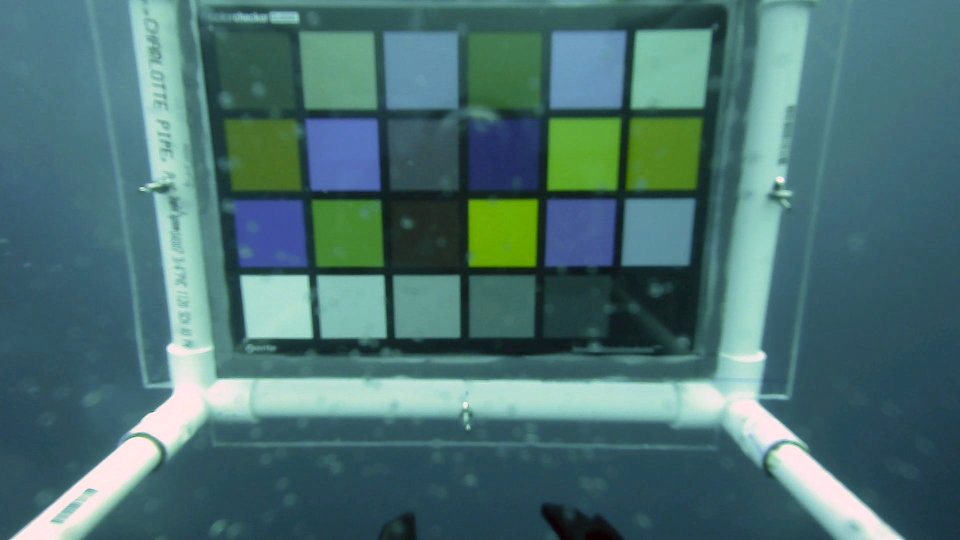}
            & \includegraphics[width=.10\textwidth, valign=c]{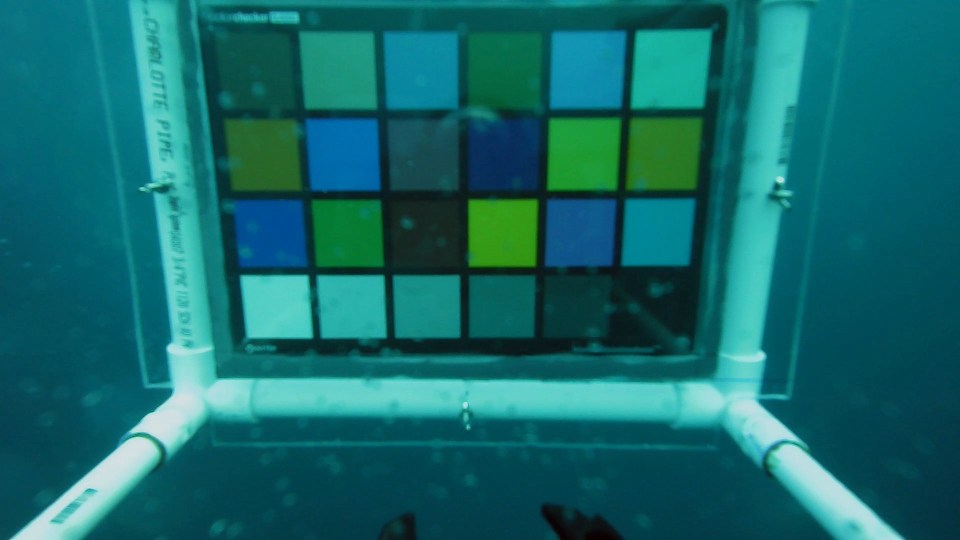}
            & \includegraphics[width=.10\textwidth, valign=c]{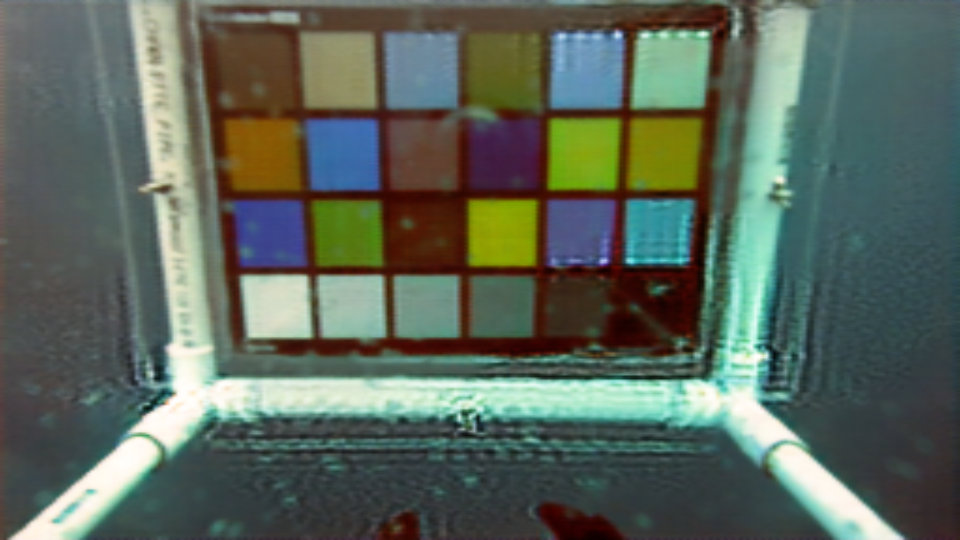}
            & \includegraphics[width=.10\textwidth, valign=c]{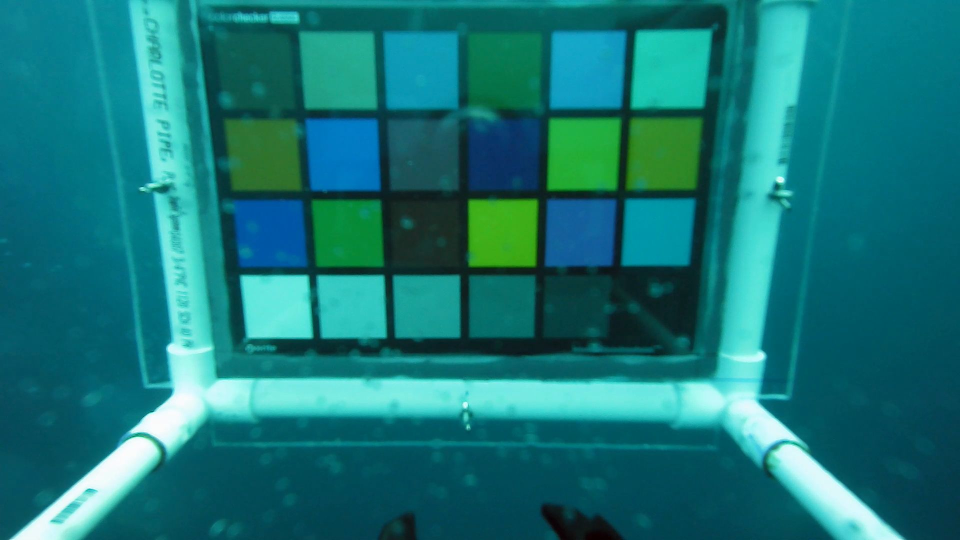}
            & \includegraphics[width=.10\textwidth, valign=c]{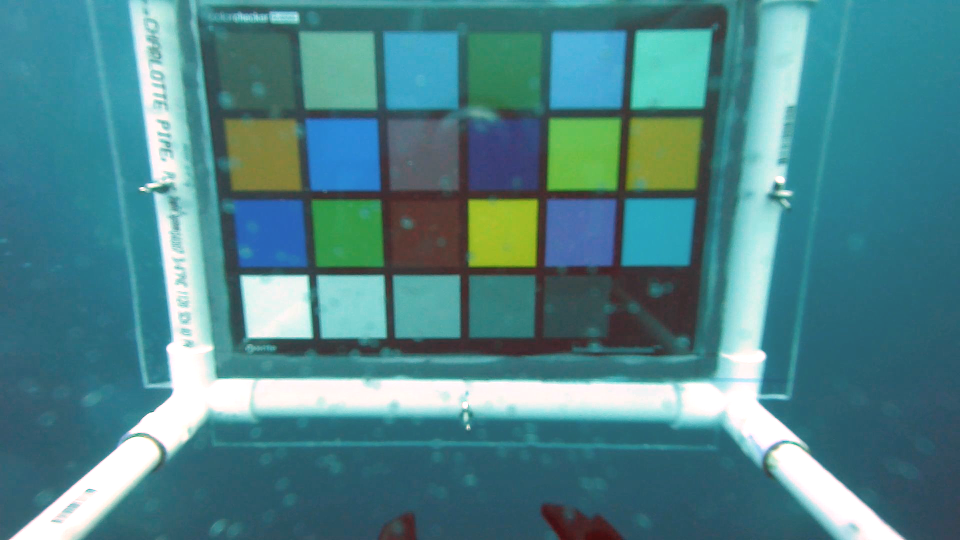} \\
    \SI{15.11}{\m} & \includegraphics[width=.10\textwidth, valign=c]{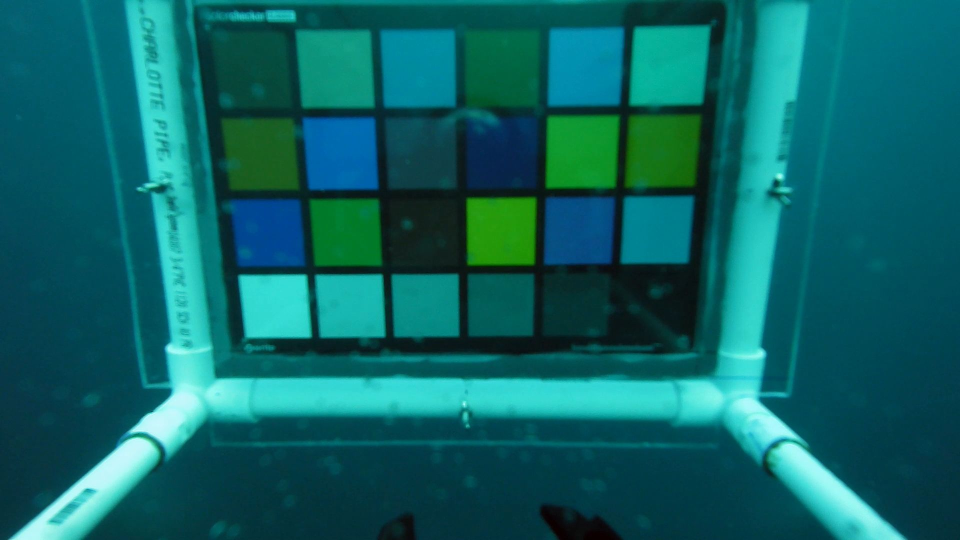}
            & \includegraphics[width=.10\textwidth, valign=c]{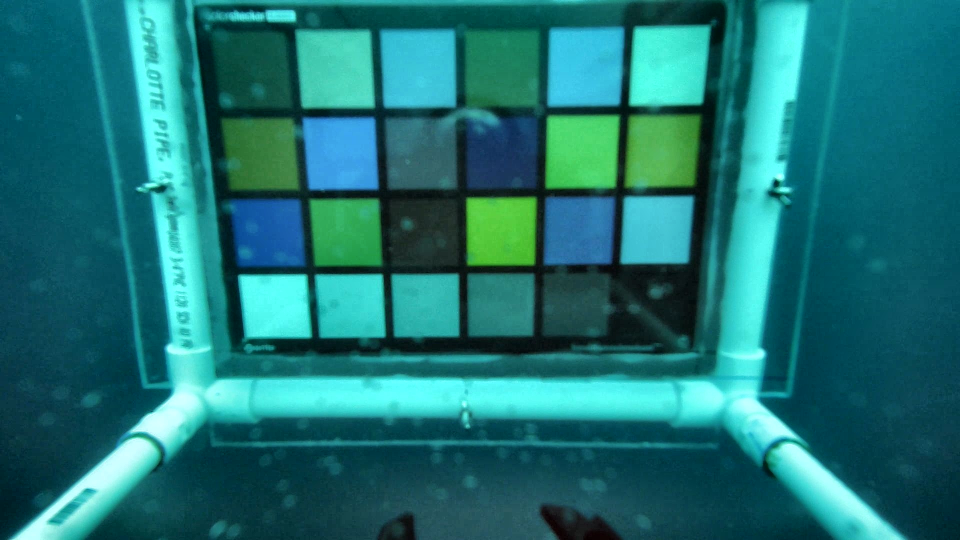} 
            & \includegraphics[width=.10\textwidth, valign=c]{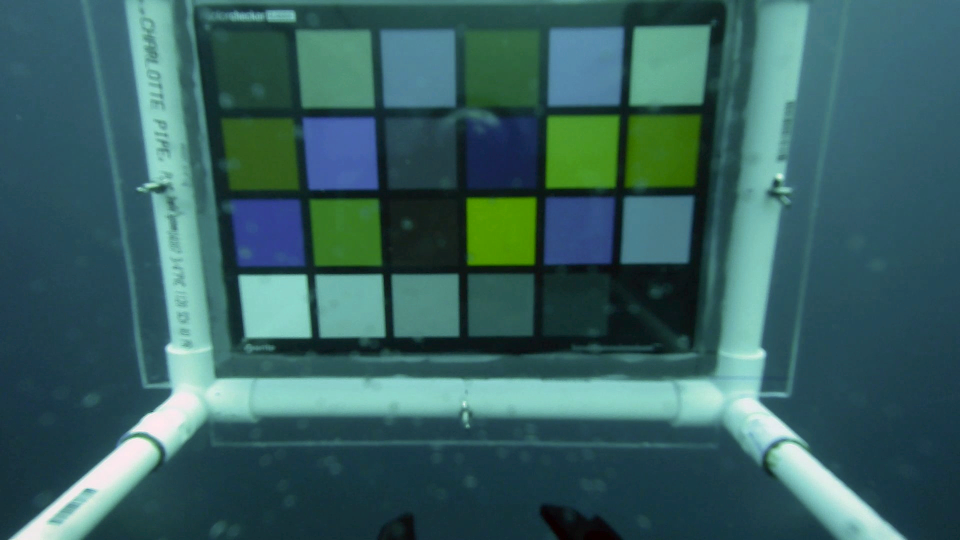}
            & \includegraphics[width=.10\textwidth, valign=c]{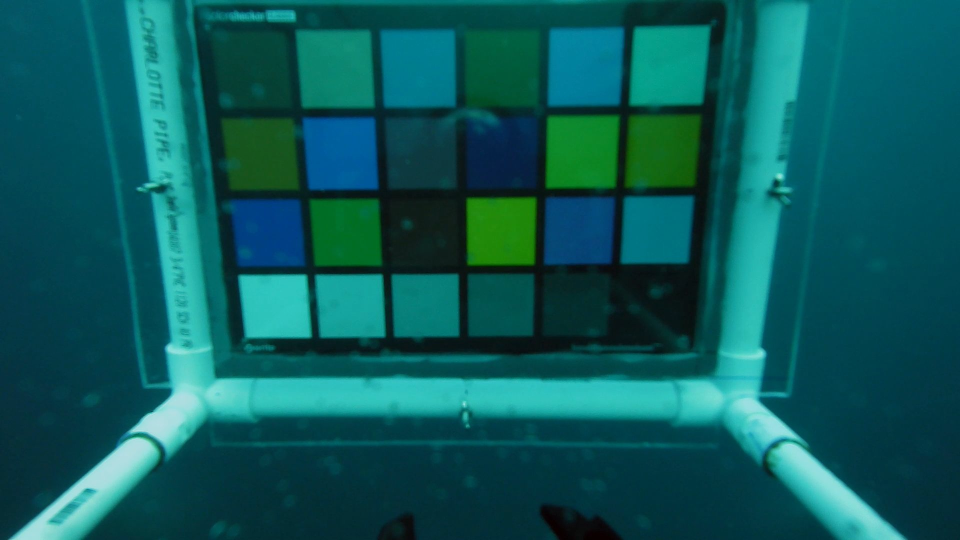}
            & \includegraphics[width=.10\textwidth, valign=c]{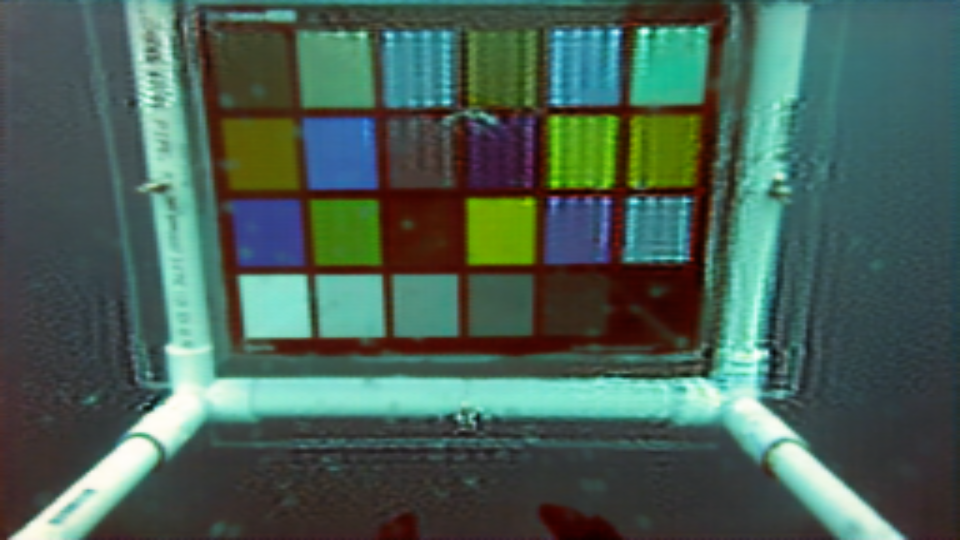}
            & \includegraphics[width=.10\textwidth, valign=c]{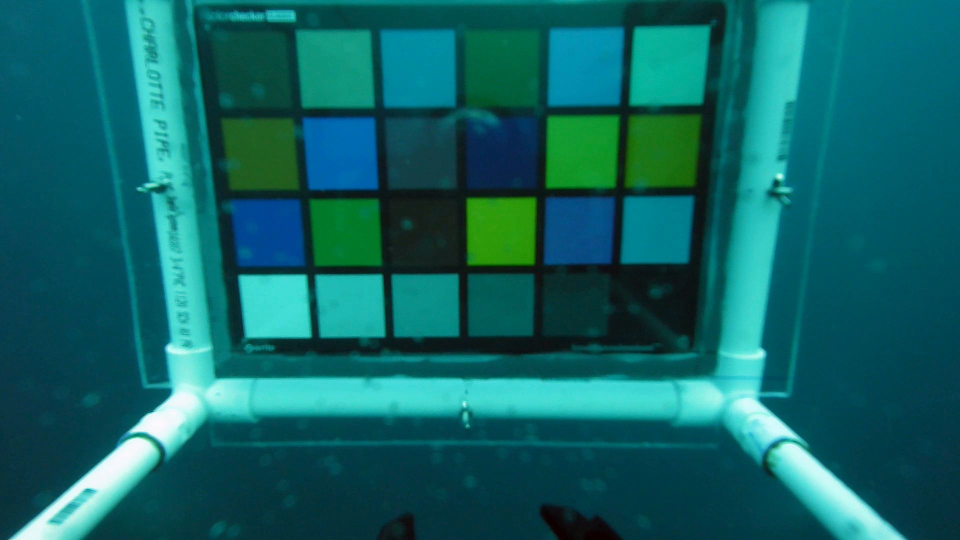}
            & \includegraphics[width=.10\textwidth, valign=c]{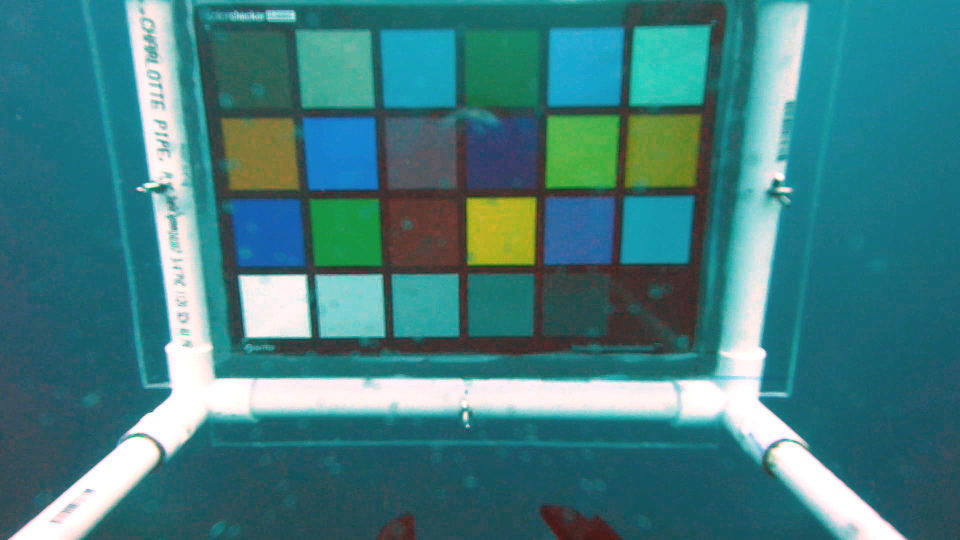} \\
    \bottomrule
  \end{tabular}
  \caption{Table of figures. \textbf{a:} Images taken by the BlueROV2 in the Caribbean Sea. \textbf{b:} Histogram equalization. \textbf{c:} White-balancing using gray world assumption. \textbf{d:} Current underwater image formation model. \textbf{e:} UGAN~\cite{fabbri2018enhancing} \textbf{f:} Fusion method~\cite{ancuti2012enhancing}. \textbf{g:} Proposed method (Est. Att.).}
  \label{tbl:table_of_corrected_images}
\end{table*}

Our first experiment measures the color accuracy of color correcting images that were taken at different depths and viewing distances. We used the X-Rite ColorChecker\textsuperscript{\tiny\textregistered} Classic as our color reference and attached it to the BlueROV2 at set distances from the camera, as depicted in Figure \ref{fig:robot-with-color-chart}. For quantitative validation, color accuracy is measured by normalized distance measurements, while color consistency is measured by variance and Euclidean mean measurements.

For measuring color changes in depth, we deployed the BlueROV2 and recorded its images as it descended down to about \SI{25}{\m}. The color chart attached to the BlueROV2 was set at a distance of \SI{0.33}{\m} from the camera. Table \ref{tbl:table_of_corrected_images} displays the images that were taken at recorded depths of \SI{3.26}{\m}, \SI{6.06}{\m}, \SI{8.98}{\m}, \SI{12.25}{\m}, and \SI{15.11}{\m}, as well as the results of applying different image color correction methods.

We ran the experiment with CLAHE, white-balancing using gray world assumption, current underwater image formation model (CUIFM), UGAN~\cite{fabbri2018enhancing}, Fusion method~\cite{ancuti2012enhancing}, and our proposed method. We assume the environment to be of Jerlov IA \cite{Solonenko2015}, for calculating the veiling light.

We implement two quantitative metrics to assess the color accuracy and consistency of the applied color correction methods. Color accuracy was tested on the \SI{15.11}{\m} depth images and represented by the Euclidean distance of intensity-normalized color for six of the color patches to the ground truth. The results shown in Table \ref{tbl:table_of_color_accuracy} present that our proposed method has the highest accuracy for red and yellow patches, making it able to restore red color, which is the first one to be almost completely absorbed at \SI{5}{\m}. One reason for less accuracy in blue and green patches is that the method must compensate for very small red channel values.

Color consistency was evaluated based on the color value changes of the blue, green, and red color patches on the color chart over depth. The variance was calculated using color samples from the \SI{3.26}{\m}, \SI{6.06}{\m}, \SI{8.98}{\m}, \SI{12.25}{\m}, and \SI{15.11}{\m} depth images. The Euclidean distance of the mean color value to the ground truth was also calculated. The results are displayed in \tab{tbl:table_of_color_consistency}. While the proposed method has high variance, it also has the lowest mean Euclidean distance error, in other words, the highest mean color accuracy. As a note, some methods present low variance, however the mean color value is far from the ground truth color. If the estimated attenuation values in the proposed method are optimized with non-linear least squares over the ground truth values and raw values, the variance decreases with a slight decrease of the mean color accuracy, as shown in the last row of the table.

We also test the color chart at different distances from the BlueROV2 camera, at a depth close to \SI{2}{\m}. Table \ref{tbl:table_of_change_in_dist_images} displays the images of the color chart set at two distances, \SI{0.33}{\m} and \SI{0.98}{\m}. As water conditions were poor, causing an increase in particle concentration, we assumed the veiling light to be the background color. We performed the experiment using UGAN~\cite{fabbri2018enhancing} and Fusion method~\cite{ancuti2012enhancing}. The results illustrate that the proposed method is also exceptional at enhancing the color over distance.

The results of this set of experiments demonstrate the high color accuracy and adequate color consistency produced from the proposed method. By sampling color from the image, the proposed method accurately estimates the attenuation attributes that best enhances the color in the images taken by the robot. Next, we show the first steps on incorporating a SLAM application to the image enhancement process.

\begin{table}
\begin{center}
\scalebox{0.75}{
\begin{tabular}{ |c|c|c|c|c|c|c|c|c| } 
 \hline
 \textbf{Methods} & \textbf{Blue} & \textbf{Green} & \textbf{Red} & \textbf{Yellow} & \textbf{Magenta} & \textbf{Cyan} \\ 
 \hline
 Raw & 0.22250 & 0.29992 & 0.64118 & 0.79897 & 0.72249 & 0.07800 \\ 
 \hline
 CLAHE & 0.19317 & \textbf{0.13941} & 0.56038 & 0.58671 & 0.57944 & 0.39448\\
 \hline
 Gray World & \textbf{0.03767} & 0.16710 & 0.62410 & 0.64116 & 0.54180 & 0.27699\\
 \hline
 CUIFM & 0.22869 & 0.33328 & 0.65016 & 0.82658 & 0.72871 & 0.14451 \\
 \hline
 UGAN~\cite{fabbri2018enhancing} & 0.05142 & 0.24070 & 0.47872 & 0.52557 & \textbf{0.49610} & 0.22867 \\ 
 \hline
 Fusion~\cite{ancuti2012enhancing} & 0.22493 & 0.30550 & 0.62816 & 0.76944 & 0.71009 & \textbf{0.07110}\\
 \hline
 \multirow{2}{8em}{Proposed method (Est. Att.)} & 0.21970 & 0.19269 & \textbf{0.32808} & \textbf{0.23409} & 0.55949 & 0.10560\\
        & & & & & & \\
 \hline
 \multirow{2}{8em}{Proposed method (Opt. Att.)} & 0.23455 & 0.19142 & 0.38265 & 0.38577 & 0.57654 & 0.10729\\
        & & & & & & \\
 \hline
\end{tabular}
}
\caption{Normalized color distance between the true color and the color in the image at \SI{15.11}{\m} for 6 squares.}
\label{tbl:table_of_color_accuracy}
\end{center}
\end{table}

\begin{table}
\begin{center}
\scalebox{0.9}{
\begin{tabular}{ |c|c|c|c|c|c| } 
 \hline
  \textbf{Methods} &  & \textbf{Blue} & \textbf{Green} & \textbf{Red} & \textbf{Average} \\ 
 \hline
 \multirow{2}{7em}{Raw} & Variance & 0.1955 & 0.2737 & 0.6795 & \textit{0.3829} \\
                   & Mean Err. & 0.19856 & 0.25588 & 0.52021 & \textit{0.32489} \\
 \hline
 \multirow{2}{7em}{CLAHE} & Variance & 0.4403 & 0.5379 & 1.3369 & \textit{0.7717} \\
                     & Mean Err. & 0.18270 & 0.09848 & 0.45547 & \textit{0.24555} \\
 \hline
 \multirow{2}{7em}{Gray World} & Variance & 0.3435 & 0.4536 & 0.6079 & \textit{0.4683} \\
                          & Mean Err. & 0.03367 & 0.17243 & 0.50145 & \textit{0.23585} \\
 \hline
 \multirow{2}{7em}{CUIFM} & Variance & 0.1766 & 0.3199 & 0.5573 & \textit{0.3513} \\
                           & Mean Err. & 0.20178 & 0.29027 & 0.54152 & \textit{0.34452} \\
 \hline
 \multirow{2}{7em}{UGAN \cite{fabbri2018enhancing}} & Variance & 0.87143 & 0.93104 & 0.71207 & \textit{0.83818} \\
                        & Mean Err. & 0.29635 & 0.31100 & 0.50722 & \textit{0.37152} \\
 \hline
 \multirow{2}{7em}{Fusion~\cite{ancuti2012enhancing}} & Variance & 0.2618 & 0.3160 & 0.2618 & \textit{0.2799} \\
                      & Mean Err. & 0.20249 & 0.26840 & 0.51964 & \textit{0.33018} \\
 \hline
 \multirow{2}{7em}{Prop. method (Est. Att.)} & Variance & 0.5181 & 0.9115 & 0.5464 & \textit{0.6587} \\
                         & Mean Err. & 0.18707 & 0.11740 & 0.21073 & \textit{0.17173} \\
 \hline
 \multirow{2}{7em}{Prop. method (Opt. Att.)} & Variance & 0.1840 & 0.3199 & 0.4093 & \textit{0.3044} \\
                         & Mean Err. & 0.22171 & 0.14167 & 0.28134 & \textit{0.21490} \\
 \hline
\end{tabular}
}
\caption{Variance of the pixel color over depth of the corrected image corresponding to each of the three color squares. Also, mean Euclidean distance error from the ground truth.}
\label{tbl:table_of_color_consistency}
\end{center}
\end{table}

\begin{table}
  \centering
  \begin{tabular}{c c c}
    \toprule
     & \textbf{0.33 m} & \textbf{0.98 m} \\
    \midrule
    a & \includegraphics[width=.15\textwidth, valign=c]{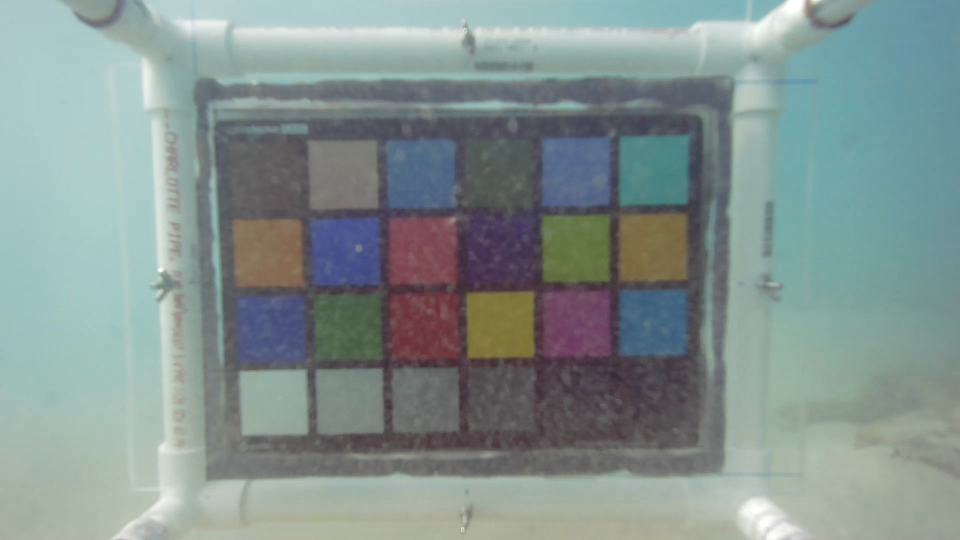}
      & \includegraphics[width=.15\textwidth, valign=c]{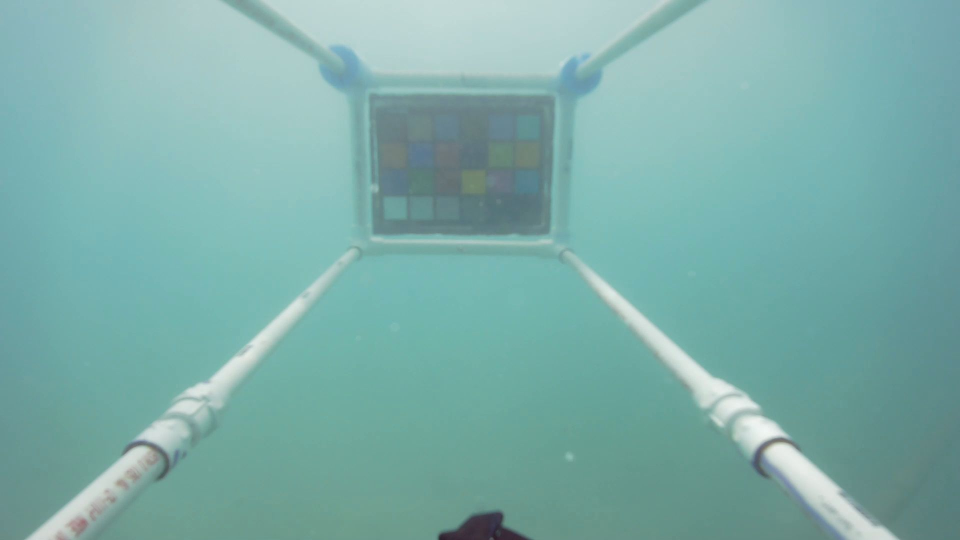} \\
    b & \includegraphics[width=.15\textwidth, valign=c]{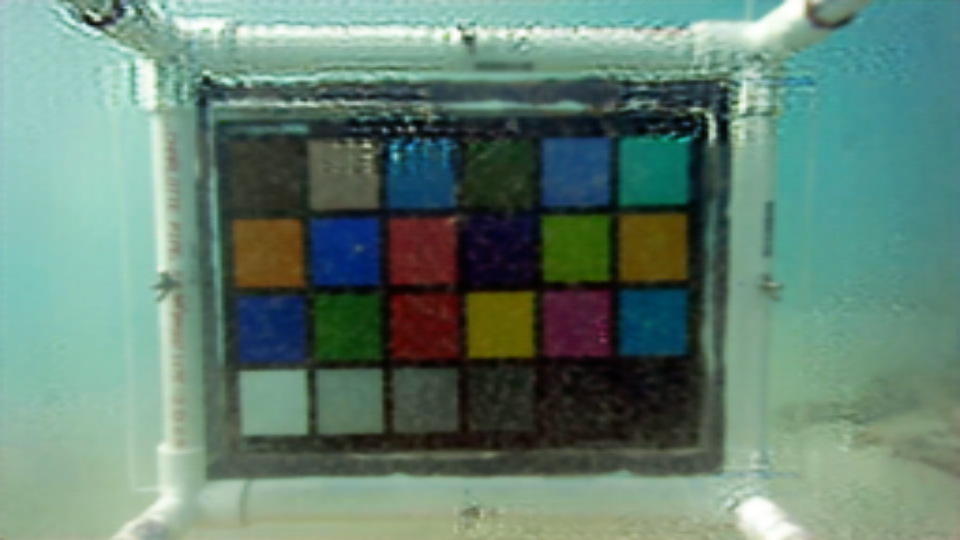}
      & \includegraphics[width=.15\textwidth, valign=c]{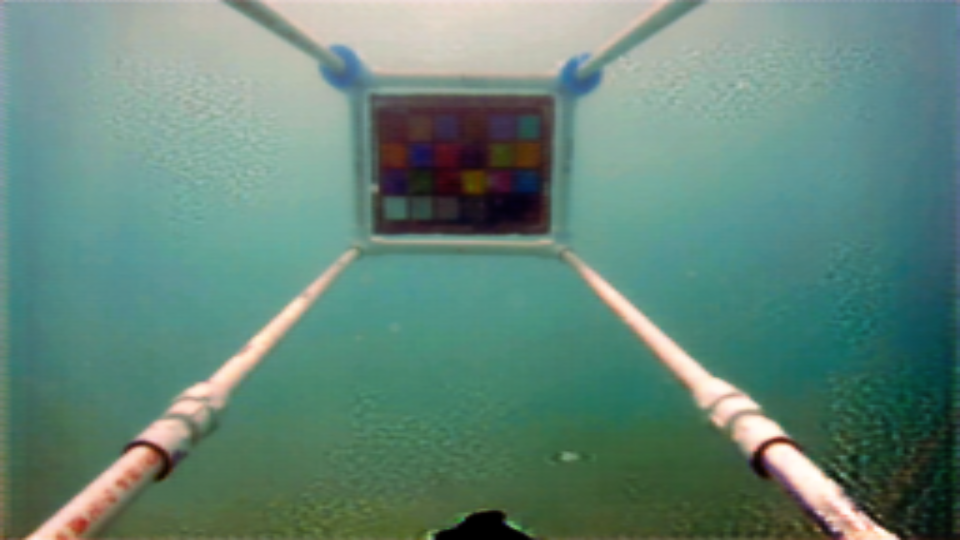} \\
    c & \includegraphics[width=.15\textwidth, valign=c]{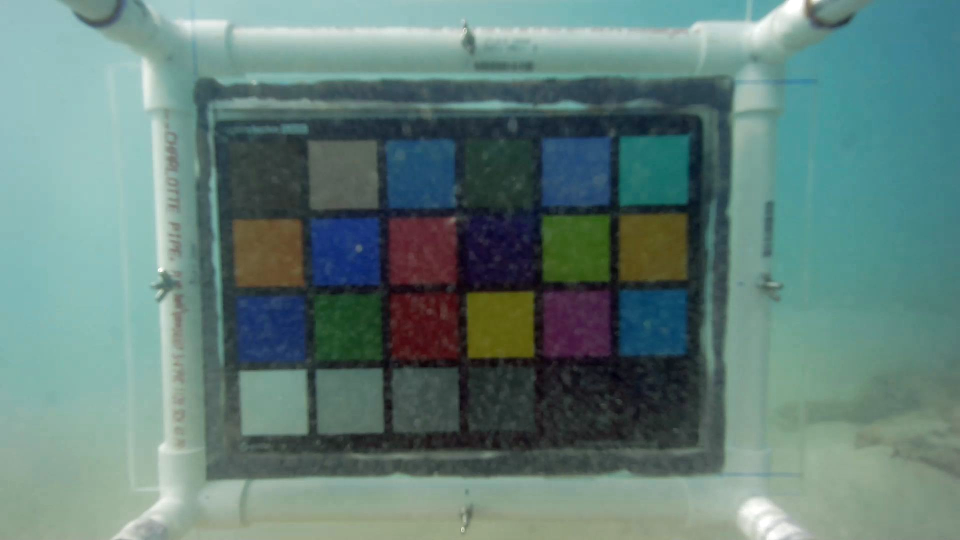}
      & \includegraphics[width=.15\textwidth, valign=c]{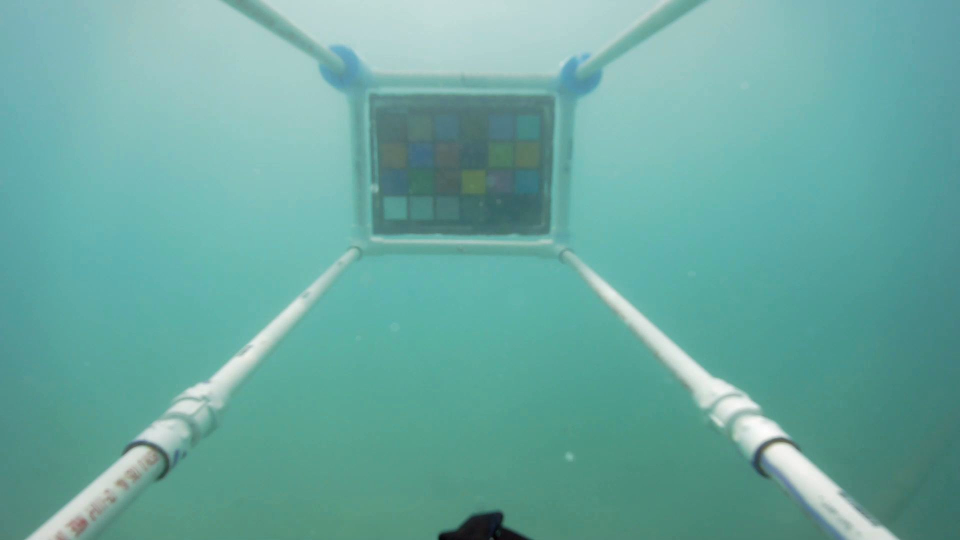} \\
    d & \includegraphics[width=.15\textwidth, valign=c]{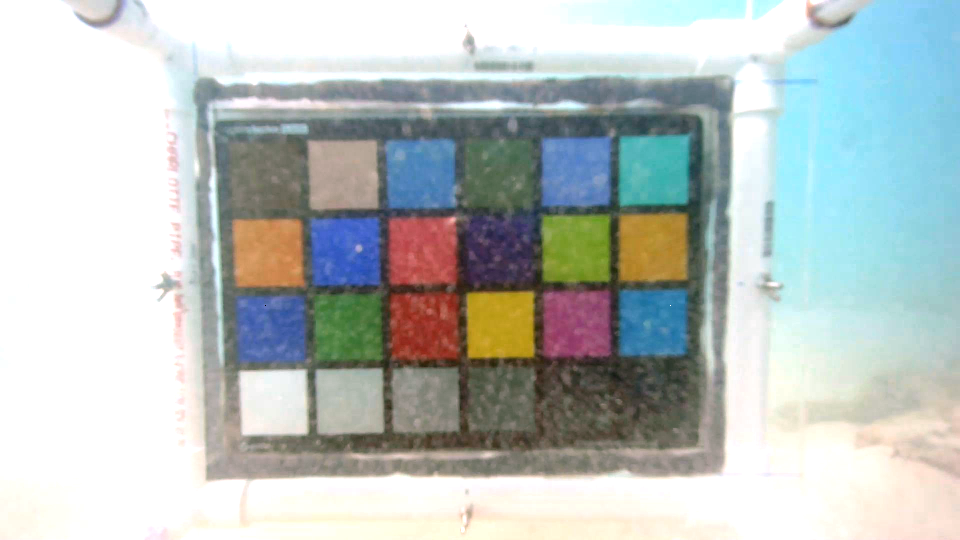}
      & \includegraphics[width=.15\textwidth, valign=c]{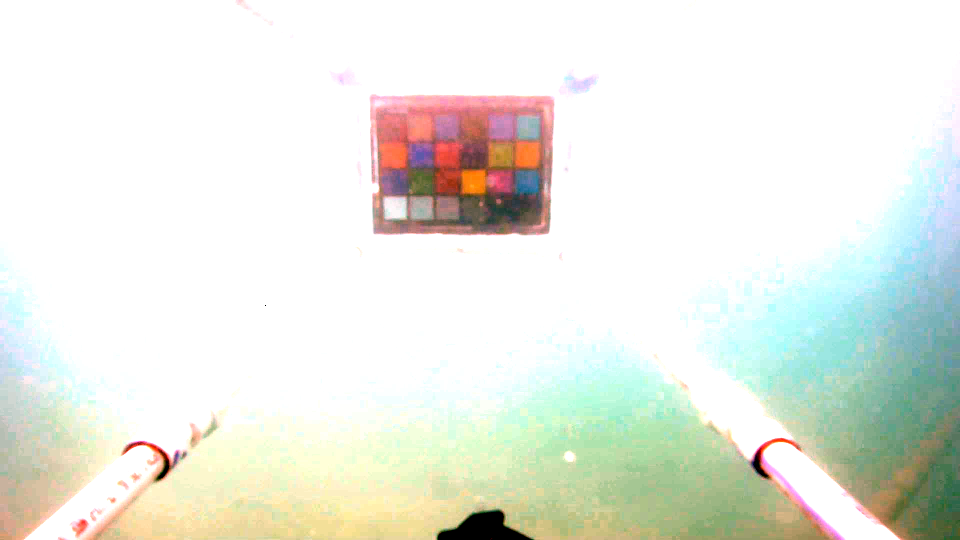} \\
    \bottomrule
  \end{tabular}
  \caption{Image enhancement over viewing distance. Images taken at a depth of around \SI{2}{\m}. Left column: color chart \SI{0.33}{\m} away. Right column: color chart \SI{0.98}{\m} away. \textbf{a:} Raw. \textbf{b:} UGAN~\cite{fabbri2018enhancing}. \textbf{c.} Fusion method~\cite{ancuti2012enhancing}. \textbf{d:} Proposed method (Est. Att).}
  \label{tbl:table_of_change_in_dist_images}
\end{table}

\begin{figure}
    \centering
    \includegraphics[width=.3\columnwidth]{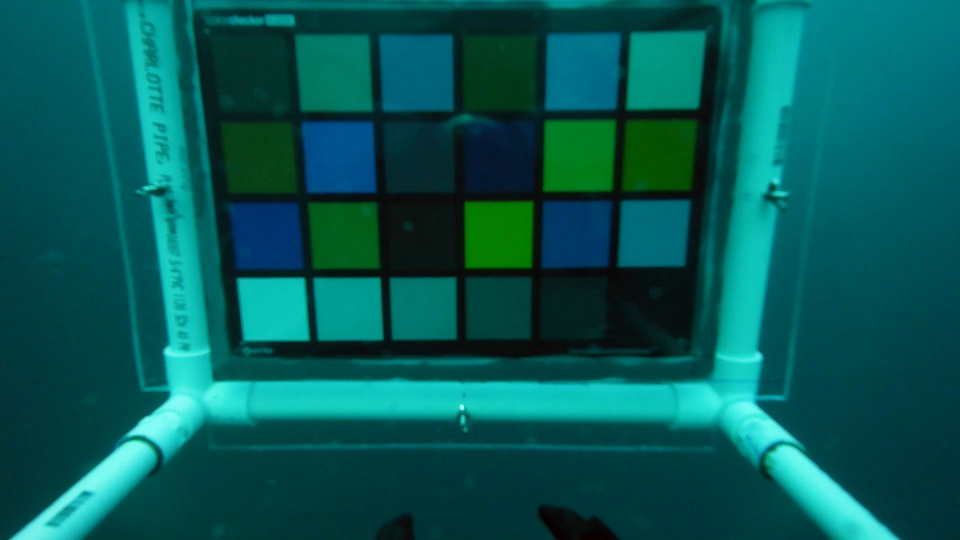}
    \includegraphics[width=.3\columnwidth]{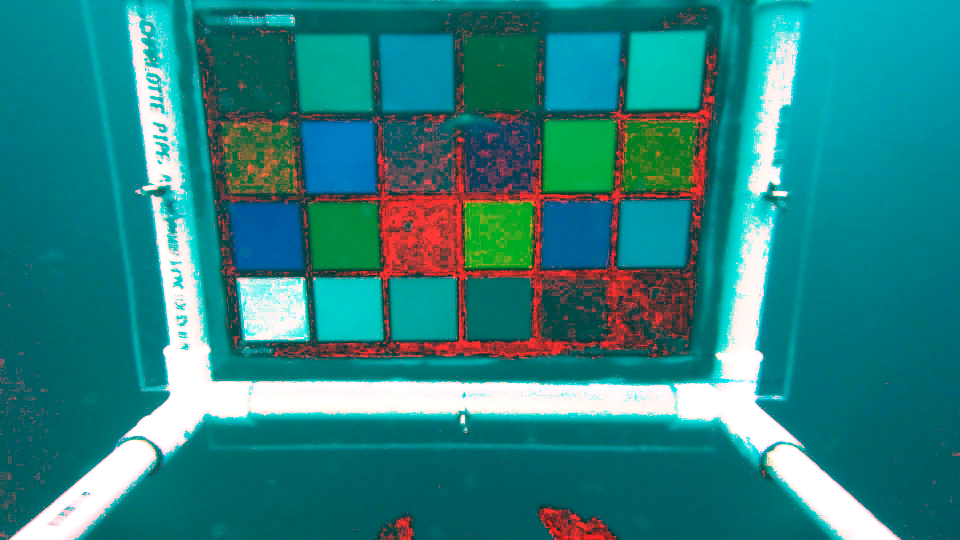}
    \includegraphics[width=.3\columnwidth]{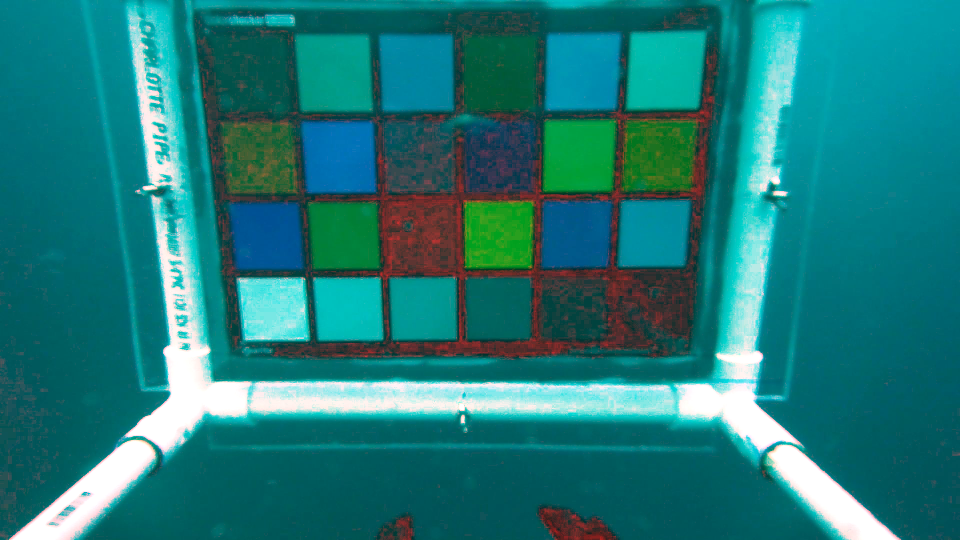}
    \caption{Image enhancement at a depth of \SI{25}{\m}. Left: Raw. Middle: Proposed method (Est. Att). Right: proposed method (Opt. Att).}
    \label{fig:25.10_depth}
\end{figure}

\begin{table}
  \centering
  \begin{tabular}{c c}
      \includegraphics[width=.45\columnwidth, valign=c]{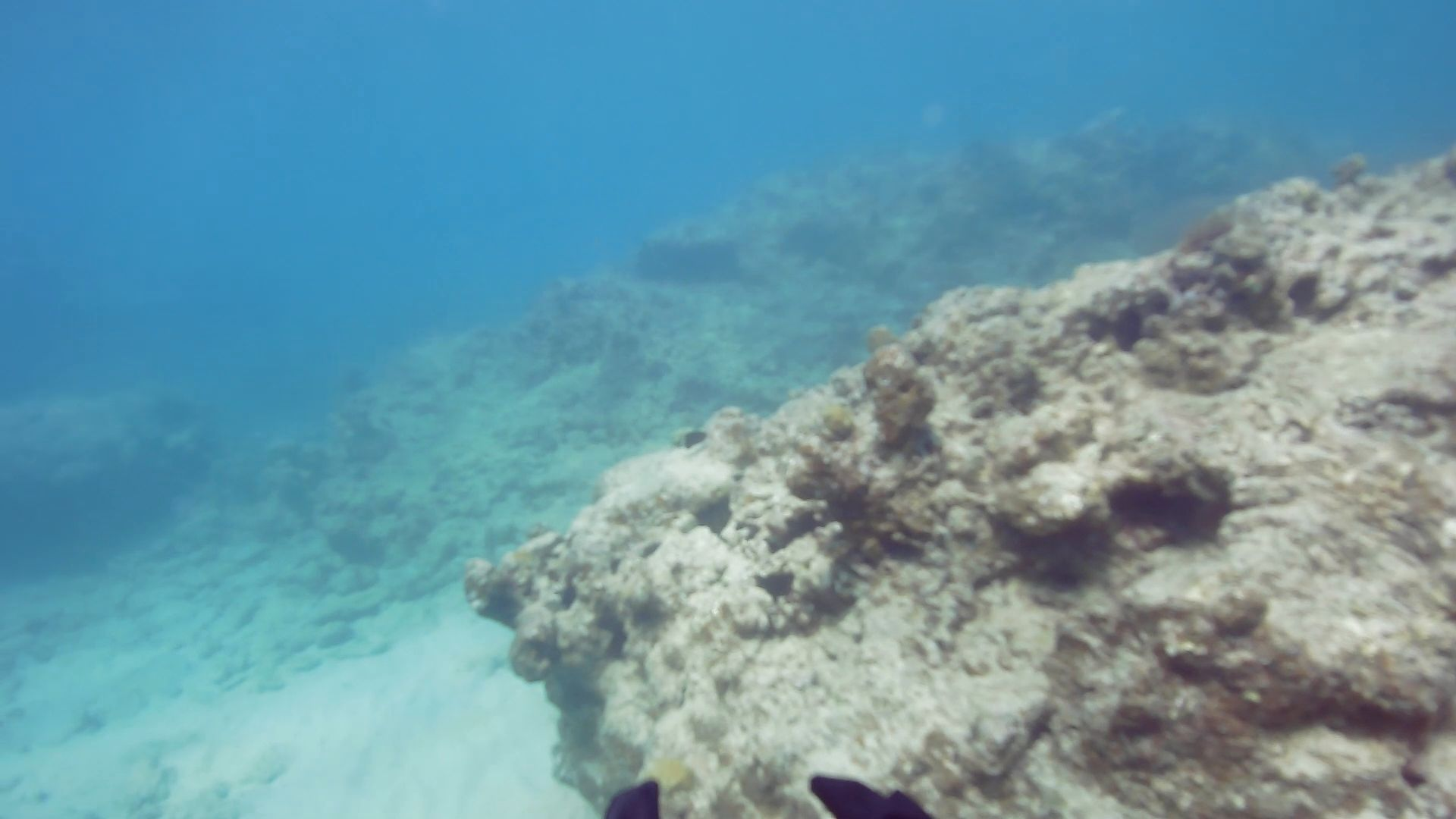}
      & \includegraphics[width=.45\columnwidth, valign=c]{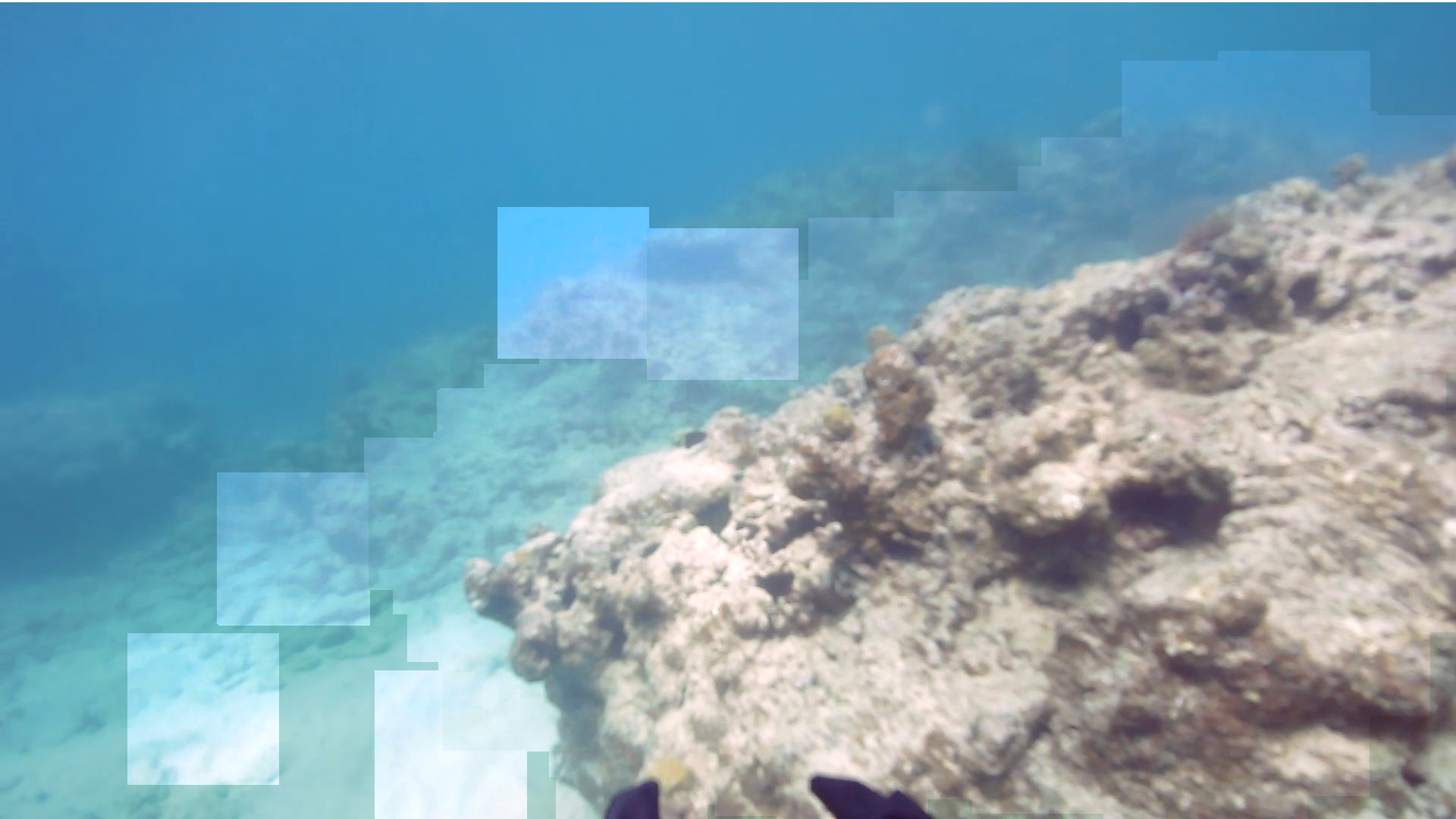} \\
      \includegraphics[width=.45\columnwidth, valign=c]{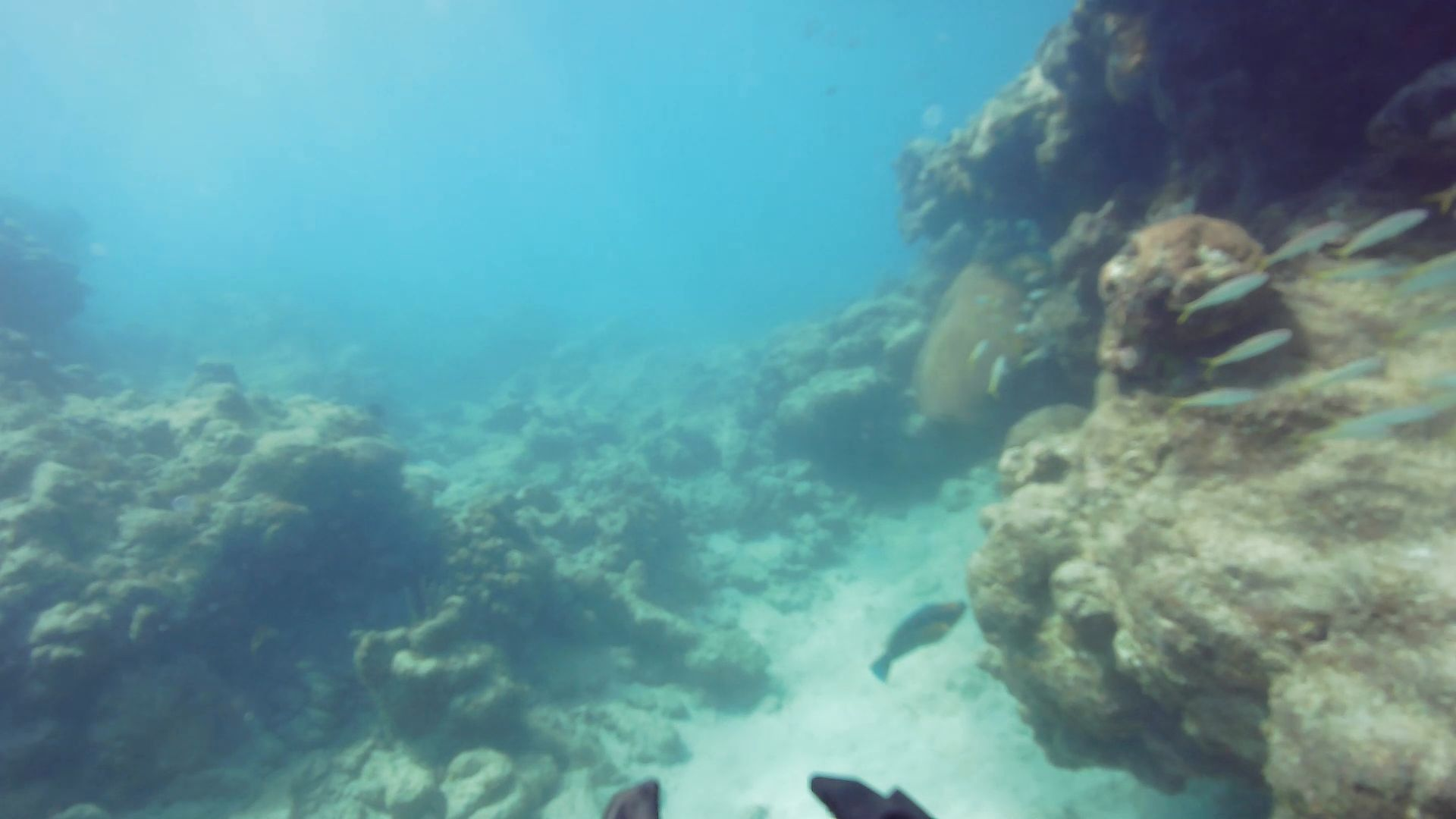}
      & \includegraphics[width=.45\columnwidth, valign=c]{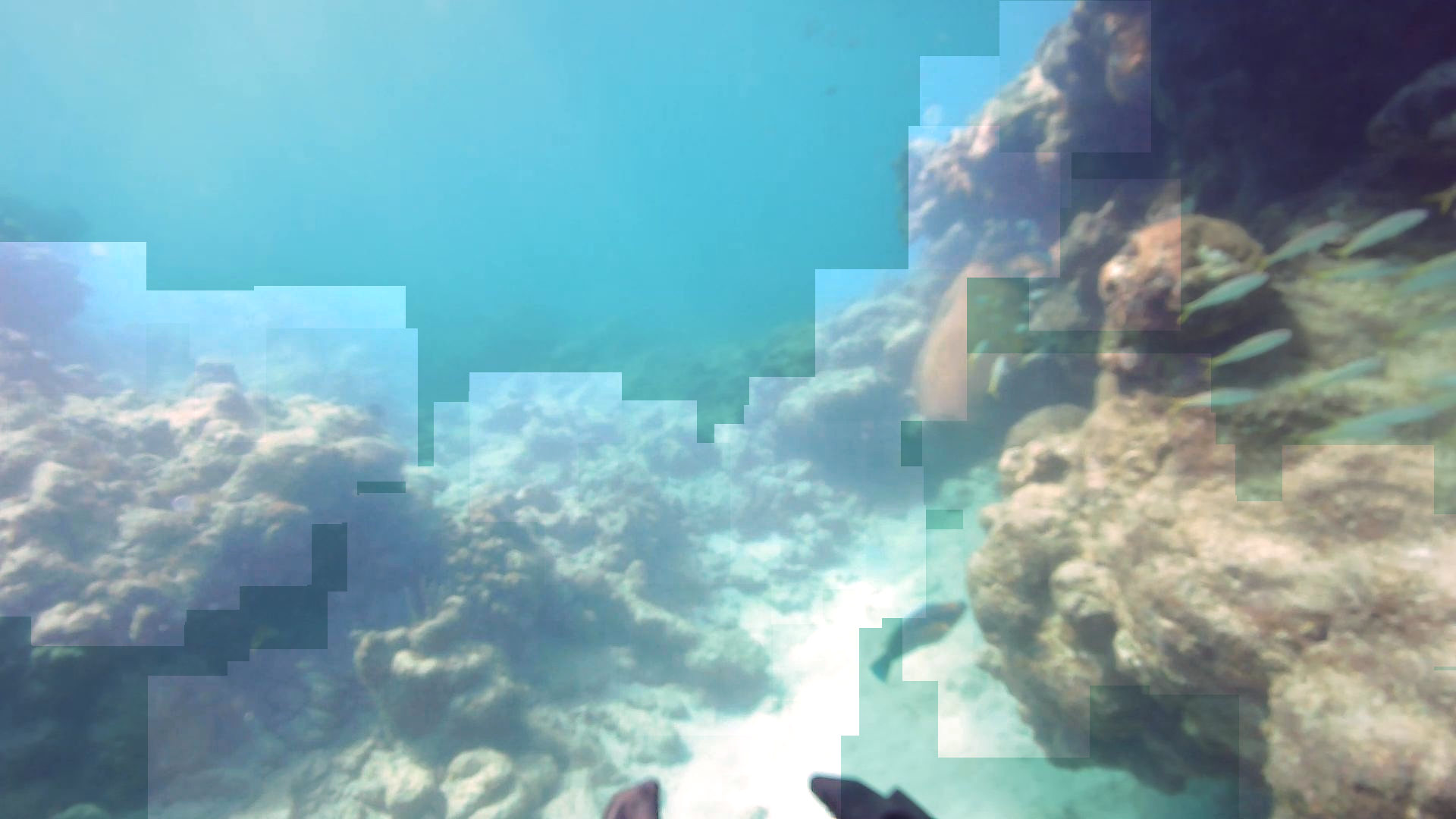} \\
      \includegraphics[width=.45\columnwidth, valign=c]{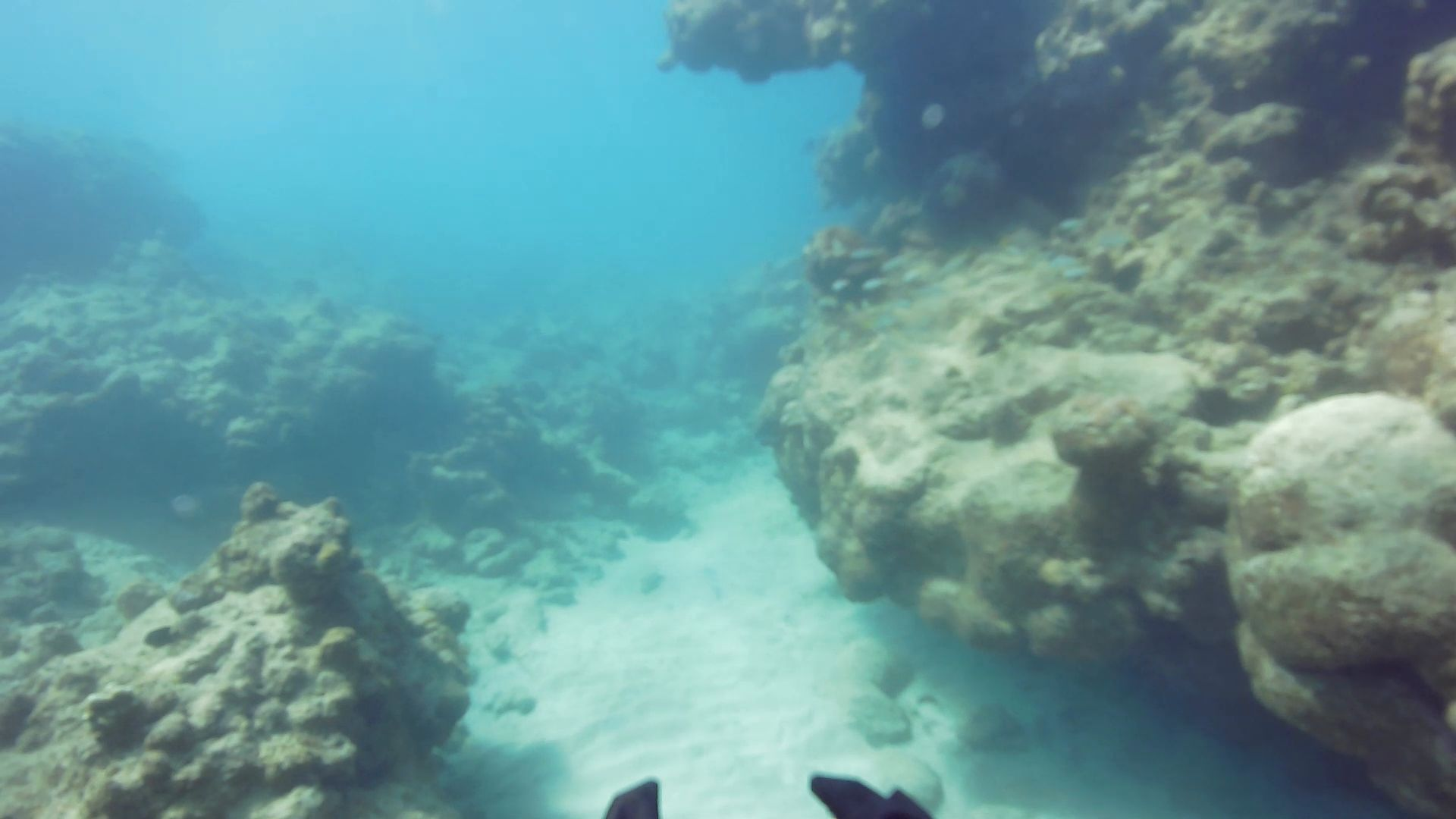}
      & \includegraphics[width=.45\columnwidth, valign=c]{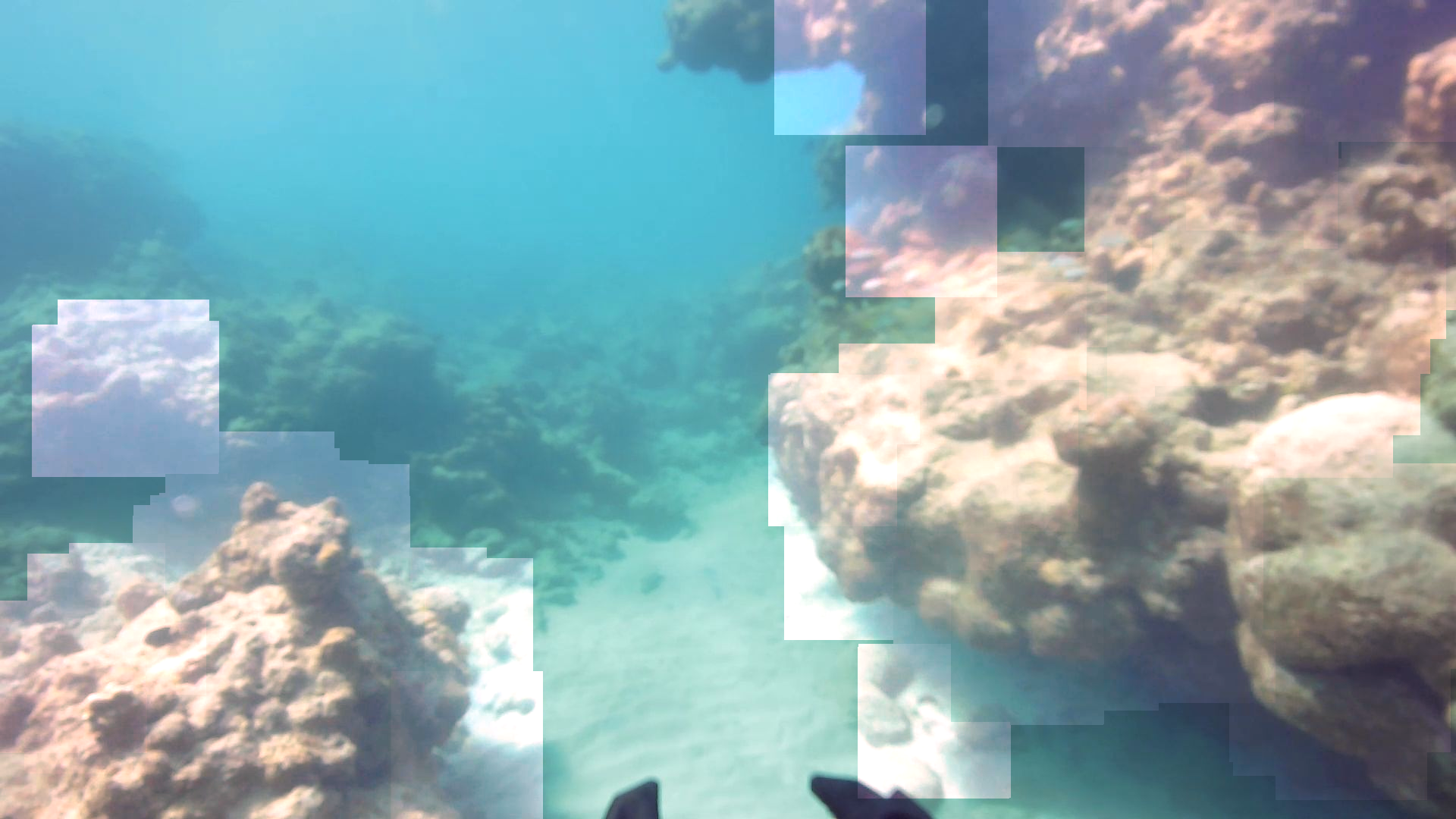} \\
      \includegraphics[width=.45\columnwidth, valign=c]{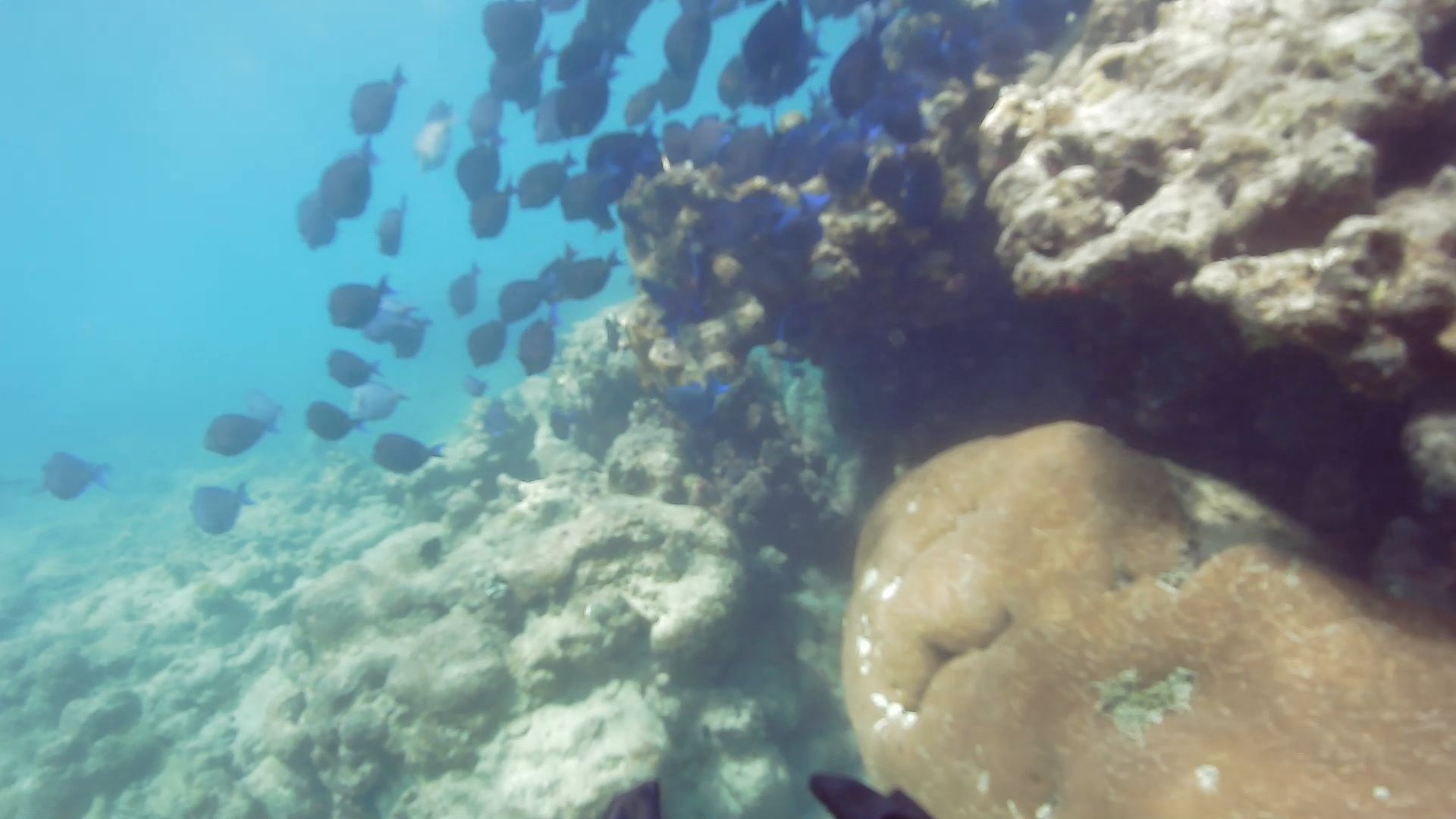}
      & \includegraphics[width=.45\columnwidth, valign=c]{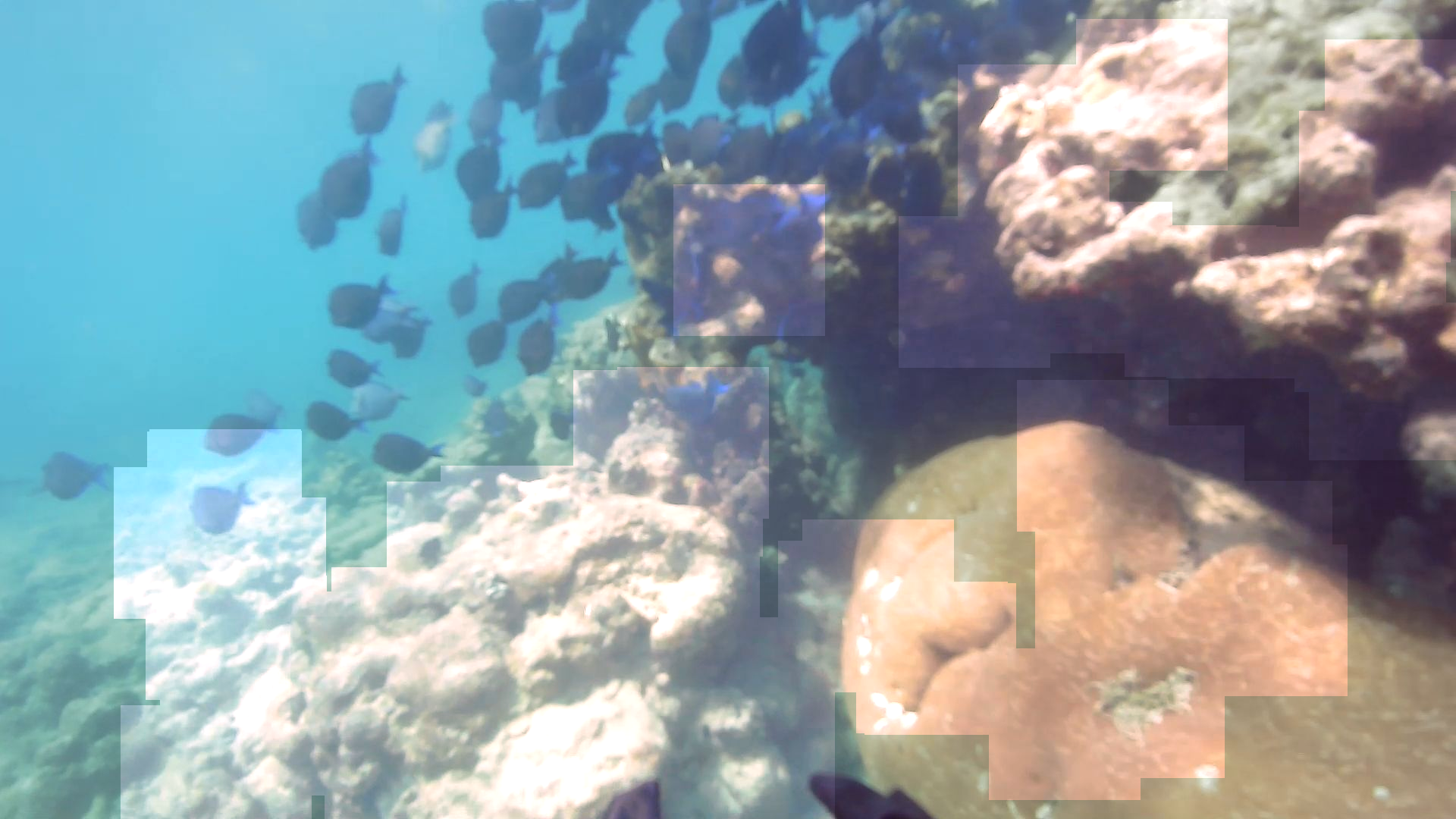} \\
      \includegraphics[width=.45\columnwidth, valign=c]{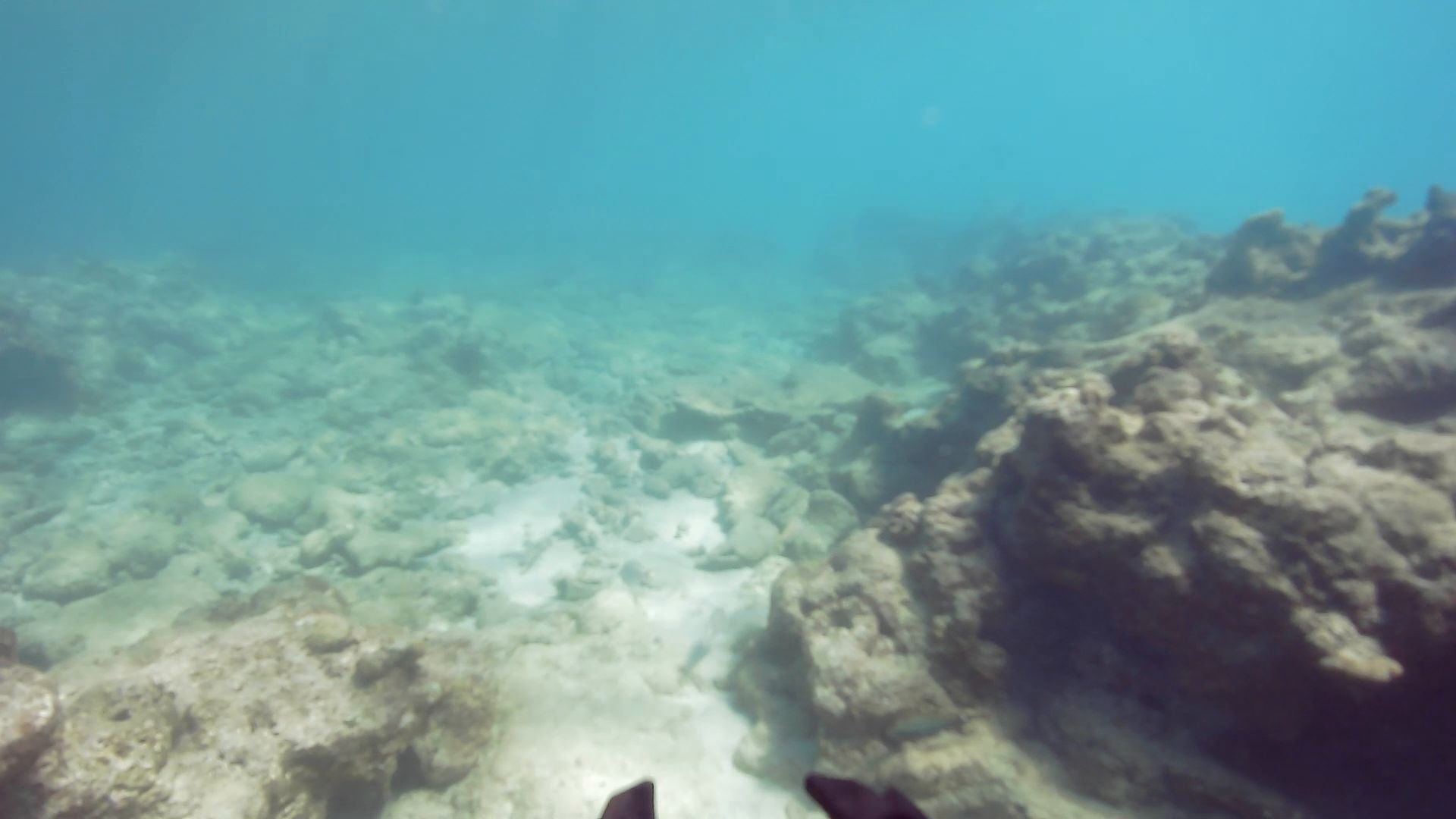}
      & \includegraphics[width=.45\columnwidth, valign=c]{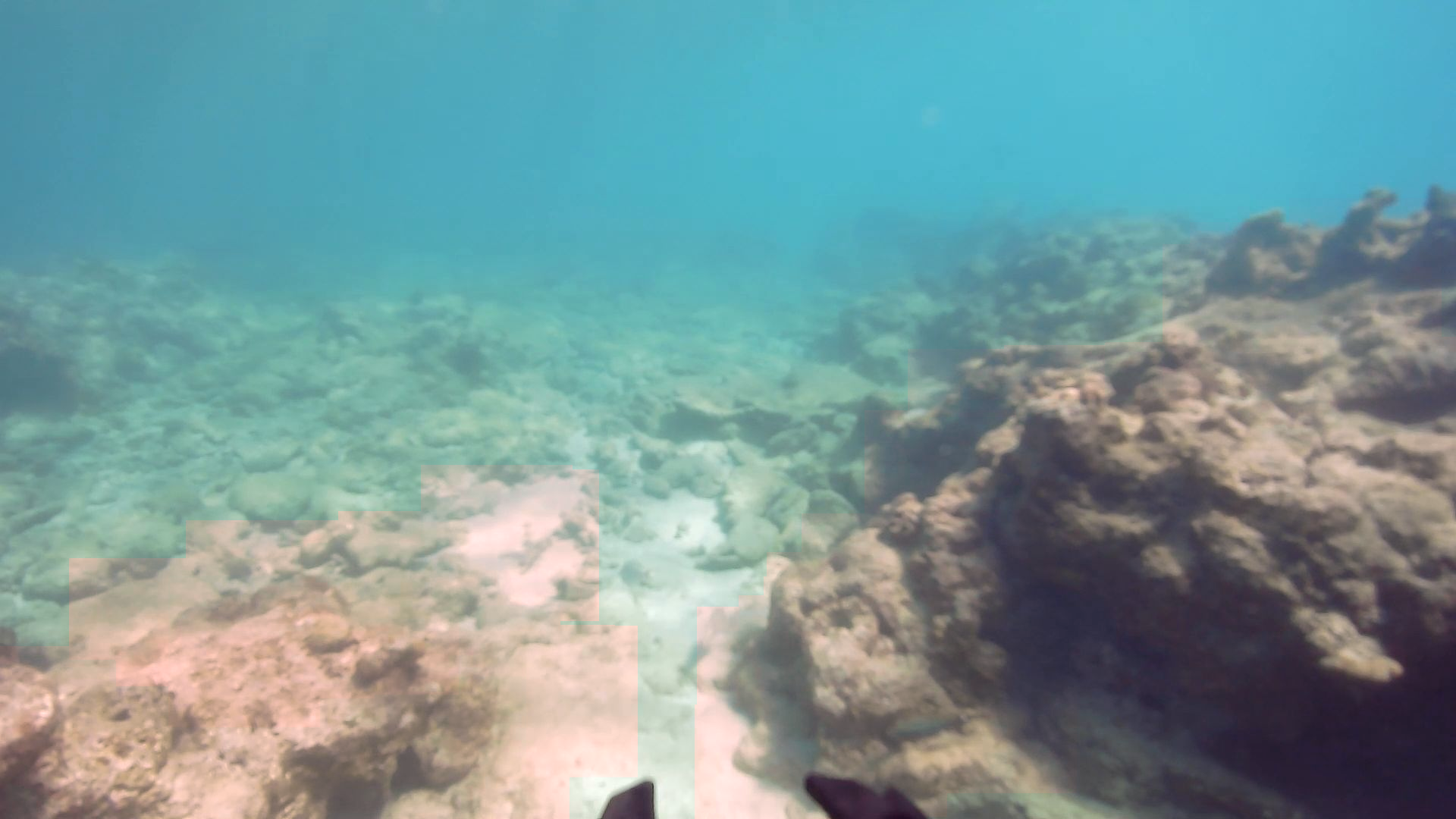} \\
  \end{tabular}
  \caption{ORB SLAM2 \cite{Mur2015} and image enhancement. Left: Raw. Right: proposed method (Opt. Att), where patches in the image are color corrected according to the tracked keypoints.}
  \label{tbl:table_of_orb_slam_implementation}
\end{table}

\subsection{SLAM Implementation}

In our second set of experiments, we demonstrate the capability of implementing robotic functions along with the proposed method. This task was the main motivator for the creation of the proposed real-time image color correction method. As a simple example, the robot was given an assignment to swim through a reef and attempt to color correct nearby areas that are tracked by a SLAM process.

We use the open-source ORB SLAM2 \cite{Mur2015} and the slightly modified Monocular ROS implementation to retrieve estimated distances of tracked points. The distances provided now are varied compared to the previous experiments. For color consistency and reducing computational power, the proposed method color corrects a fixed-size patch around each of the tracked points. Furthermore, the patches are color corrected using calculated attenuation values from the previous depth experiment using a color chart.

Table \ref{tbl:table_of_orb_slam_implementation} displays the experimental image results. The proposed method retrieves the tracked points from ORB SLAM2 and color corrects the pixels in the patches that surround these tracked points, enhancing the color and increasing the contrast between background and foreground. The overall process of image enhancement with the current implementation takes about 0.02 s. One limitation of our proposed method is that the distance values estimated from ORB SLAM2 \cite{Mur2015} are calculated using monocular assumptions. Thus, the distance values do not have the proper unit scale. Robots with stereo vision can overcome this limitation, while BlueROV2 cannot with only one camera installed.

This experiment shows the compatibility of implementing the proposed method with a SLAM application. The proposed method can be extended to other robotic applications that require good quality images.

\section{DISCUSSION}\label{sec:discussion}

The proposed method is based on a new underwater imaging formation model. Accordingly, there are calculations that are dependent on coefficients provided beforehand. For calculating the veiling light, we use the values from \cite{Solonenko2015} that are for oceanic water types. They are not meant for fresh water application, as light attenuation changes dramatically across the year, and each year might present different optical properties. As fresh water test cases were not performed, we are unsure of the effectiveness of our proposed method in these environments, and it will be our next investigation.

It was observed that at depths \SI{15}{\m} and greater the value of the red channel at each pixel is close to 0, leading to a loss of information. This darkroom effect causes colors with similar channel values to be indistinguishable. Additionally, this leads to near zero division in the proposed method, which overexposes the red channel in the corrected image, as shown in Figure \ref{fig:25.10_depth}. Correcting the image using the optimized proposed method greatly removes the artifacts, but a more substantial solution will be left for future work. In future analysis, we also aim to collect more imagery data that have less noise from surrounding air bubbles, dirt, and glares.

In continuation with the ORB SLAM2 \cite{Mur2015} implementation, we would like to evaluate color correction in relation to viewing distance accuracy, as well as tracking accuracy by using enhanced images rather than raw images. One theory is that during an exploration task, an image color correction process and a SLAM implementation could work together, as in a give-and-take relationship, to increase accuracy in both of their respective tasks. We would like to see more results of our proposed method implemented in other robotic tasks.

\section{CONCLUSIONS}\label{sec:conclusion}

The underwater imaging formation model depends on multiple factors from the environment, like the water type and visibility, as well as water depth, imaging range, and camera sensor response. Our proposed method, based on this underwater imaging formation model, estimates the attenuation coefficient values based on the current image scene. Experiments demonstrate that our proposed method produces high color accuracy and consistency. Further tests reveal the potential of our proposed method to coordinate with other robotic tasks, such as SLAM. Our ongoing research agenda includes adapting image color correction to other robotic tasks, such as scene reconstruction and object manipulation.

\section*{ACKNOWLEDGMENT}
The authors would like to thank Jennifer Jain for the experimental support. This work is supported in part by the Dartmouth Burke Research Initiation Award.


\bibliographystyle{IEEEtran}
\bibliography{IEEEabrv,refs}

\end{document}